\documentclass[10pt,journal]{IEEEtran}
\usepackage{amsmath,amssymb,amsfonts}

\usepackage{balance}
\usepackage{subfig}
\usepackage{booktabs}
\usepackage{multirow}
\usepackage{xcolor}
\usepackage[hyphens]{url}
\usepackage{graphicx}

\usepackage[hidelinks]{hyperref}
\hypersetup{
pdfauthor={P. Drozdowski, F. Stockhardt, C. Rathgeb, C. Busch},
pdfsubject={Biometrics},
pdftitle={Signal-level Fusion for Indexing and Retrieval of Facial Biometric Data},
pdfkeywords={Biometric Identification, Computational Workload Reduction, Indexing, Information Fusion, Face Recognition, Morphing},
bookmarks=true,
bookmarksopen=true,
bookmarksnumbered=true,
pagebackref=false
}

\def\eg{\textit{e.g}. } 
\def\ie{\textit{i.e}. } 
\def\cf{\textit{c.f}. } 
\def\etc{\textit{etc}. } 

\def\wrt{w.r.t. } \def\etal{\textit{et al}. }

\newcommand{\set}[1]{\left\lbrace #1 \right\rbrace}

\begin{document}

\title{Signal-level Fusion for Indexing and Retrieval of Facial Biometric Data}

\author{\IEEEauthorblockN{Pawel~Drozdowski,
        Fabian~Stockhardt,
        Christian~Rathgeb,
        Christoph~Busch} \\ \vspace{0.1cm}
    \IEEEauthorblockA{da/sec -- Biometrics and Internet Security Research Group, Hochschule Darmstadt, Germany
    \\\texttt{\{name.lastname\}@h-da.de}
} \vspace{-0.5cm}
}

\maketitle

\begin{abstract}
The growing scope, scale, and number of biometric deployments around the world emphasise the need for research into technologies facilitating efficient and reliable biometric identification queries. This work presents a method of indexing biometric databases, which relies on signal-level fusion of facial images (morphing) to create a multi-stage data-structure and retrieval protocol. By successively pre-filtering the list of potential candidate identities, the proposed method makes it possible to reduce the necessary number of biometric template comparisons to complete a biometric identification transaction. The proposed method is extensively evaluated on publicly available databases using open-source and commercial off-the-shelf recognition systems. The results show that using the proposed method, the computational workload can be reduced down to around 30\%, while the biometric performance of a baseline exhaustive search-based retrieval is fully maintained, both in closed-set and open-set identification scenarios.
\end{abstract}


\section{Introduction}
\label{sec:introduction}
\begin{figure}[!ht]
\centering
\subfloat[Fingerprint]{\includegraphics[height=3.25cm]{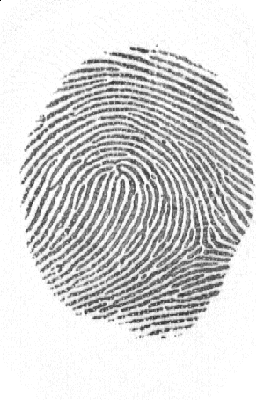}} \hfill
\subfloat[Face]{\includegraphics[height=3.25cm]{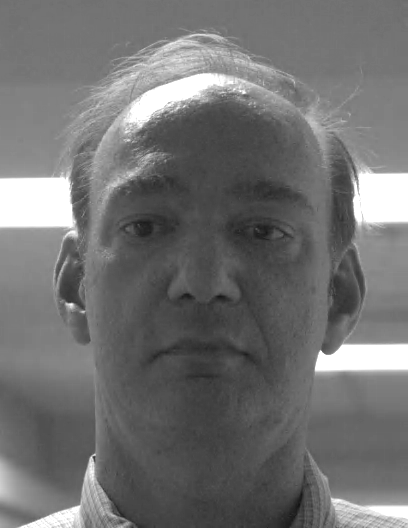}} \hfill
\subfloat[Iris]{\includegraphics[height=3.25cm]{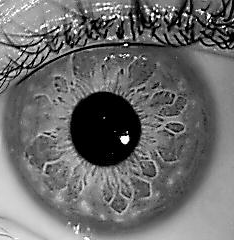}}
\caption{Example images of some commonly used biometric characteristics (from MCYT \cite{Ortega-MCYT-2003}, FRGC \cite{Phillips-FRGC-CVPR-2005}, and IITD \cite{Kumar-IITD-2010})}
\label{fig:exampleimages}
\end{figure}

Biometric technologies have become an essential component of many personal, commercial, and governmental identity management systems around the world. Both the positive (\eg access control) and the negative (\eg forensics and surveillance) identification scenario can greatly benefit through the use of biometrics. Biometrics rely on highly distinctive characteristics of human beings (see figure \ref{fig:exampleimages} for some popular examples), which make it possible for individuals to be reliably recognised using fully automated algorithms. The global market value for biometric technologies has been steadily growing in recent years and is currently estimated to be tens of billions of dollars \cite{Pascu-BiometricMarketValue-2020}. Examples of actual application scenarios of biometrics beyond personal devices (see \eg \cite{Das-MobileBiometrics-2018}) include, but are not limited to border control (see \eg \cite{EULisa-EURODAC-2016,SmartBorders-EU-2018,Gemalto-IDENT-2019,Northrop-HART-2018}), forensic investigations and law enforcement (see \eg \cite{Moses-AFIS-2010,Gemalto-AFIS-2019,FBI-CODIS-2018}), national ID systems (see \eg \cite{UIDAI-Aadhaar-2012,Dalwai-Aadhaar-2014}), as well as voter registration for elections (see \eg \cite{Bowyer-IrisElection-2015,CEPPS-Congo-2018}). Currently, the largest systems of this kind reach hundreds of millions or even over a billion enrolled subjects (see \eg \cite{UIDAI-Dashboard}).

The increasing scope, size, and prevalence of biometric systems' deployments necessitate the development of technologies capable of efficient and accurate processing of biometric data. The aforementioned systems often need to operate in biometric identification and duplicate enrolment check scenarios, where typically an exhaustive search (\ie one-to-many comparison) is needed in order to identify a potential previous enrolment based on a biometric probe. In this context, solutions which help to achieve shorter practical system response times through algorithmic methods, rather than through mere scaling of the hardware architecture are of interest; such algorithmic methods, referred to as biometric workload reduction and/or biometric indexing, can help to improve user experience and to reduce the monetary costs. The practical relevance of such methods is evidenced by a strong interest from the governmental side, manifesting itself through numerous benchmarks and competitions \cite{FRVT-Identification-2020,IREX-2018,Bust-DHSRally-2019}.

Accordingly, significant research efforts have been devoted to development of methods for computationally efficient biometric identification systems. Beyond mere software and hardware-based implementation optimisations, the idea of computational workload reduction in biometric identification has received a lot of attention from researchers. Such methods typically aim to decrease the number of template comparisons needed in a biometric identification transaction by taking advantage of certain intrinsic properties of biometric data coupled with advanced search structures and algorithms. For more details on this subject, the reader is referred to subsection \ref{subsec:background_workload} and a survey of Drozdowski \etal \cite{Drozdowski-WorkloadSurvey-IET-2019}). In contrast to methods reported in the literature, which typically rely on access to feature vectors extracted by facial recognition algorithms, this article considers the more challenging case of a black-box biometric recognition system (\ie any commercial off-the-shelf product with unknown internal representation of feature vectors in proprietary format); specifically the method proposed in this work relies merely on raw facial biometric samples and the capability to compute comparison scores between them. Furthermore, the method proposed in this work requires neither additional training nor classifiers for the purpose of computational workload reduction.

\subsection{Contribution and Organisation}
\label{subsec:introduction_organisation}
This work proposes a concept wherein signal-level information fusion is utilised as a basis for a multi-stage indexing and retrieval scheme for facial biometric data. The proposed method relies on facial image morphing (see subsection \ref{subsec:background_morphing} and a survey of Scherhag \etal \cite{Scherhag-MorphingSurvey-2019}) and filtering of candidate short-lists based on biometric comparison scores. A comprehensive experimental evaluation shows that by using the proposed concepts, the computational workload associated with a biometric identification (both closed-set and open-set) transaction can be reduced down to around 30\%, without negatively affecting the biometric performance in terms of recognition accuracy. 

By virtue of relying on signal-level fusion, the proposed concepts could be integrated even into black-box face recognition systems, \ie without accessing the underlying algorithms and feature representations. This contribution is in stark contrast to numerous other methods published in this area, as they very often rely on unrestricted access to biometric templates (\ie extracted feature sets) to facilitate indexing and efficient retrieval.

The work presented in this article significantly extends the proof-of-concept publication of this method in \cite{Drozdowski-MorphingPreselection-ICASSP-2019}. More specifically, the conceptual and technical contributions beyond the original conference paper include:

\begin{itemize}
\item New methods of intelligent pairing of parent images to be morphed, which significantly improve the originally achieved results.
\item A theoretical and empirical analysis of an extension of the originally proposed two-stage retrieval scheme to incorporate multiple stages.
\item A larger evaluation dataset.
\item Numerous additional open-source (OSS) and commercial off-the-shelf (COTS) morphing tools and face recognition systems used in the evaluation.
\end{itemize}

The remainder of this article is organised as follows: section \ref{sec:background} provides the relevant background information and surveys the related works. The proposed system is described in section \ref{sec:proposedsystem}. The experimental setup is outlined in section \ref{sec:experimentalsetup}, while the experimental results are presented in section \ref{sec:results}. Concluding remarks and a summary are given in section \ref{sec:conclusion}.

\begin{figure}[!ht]
\centering
\subfloat[Subject 1]{\includegraphics[width=0.3\columnwidth]{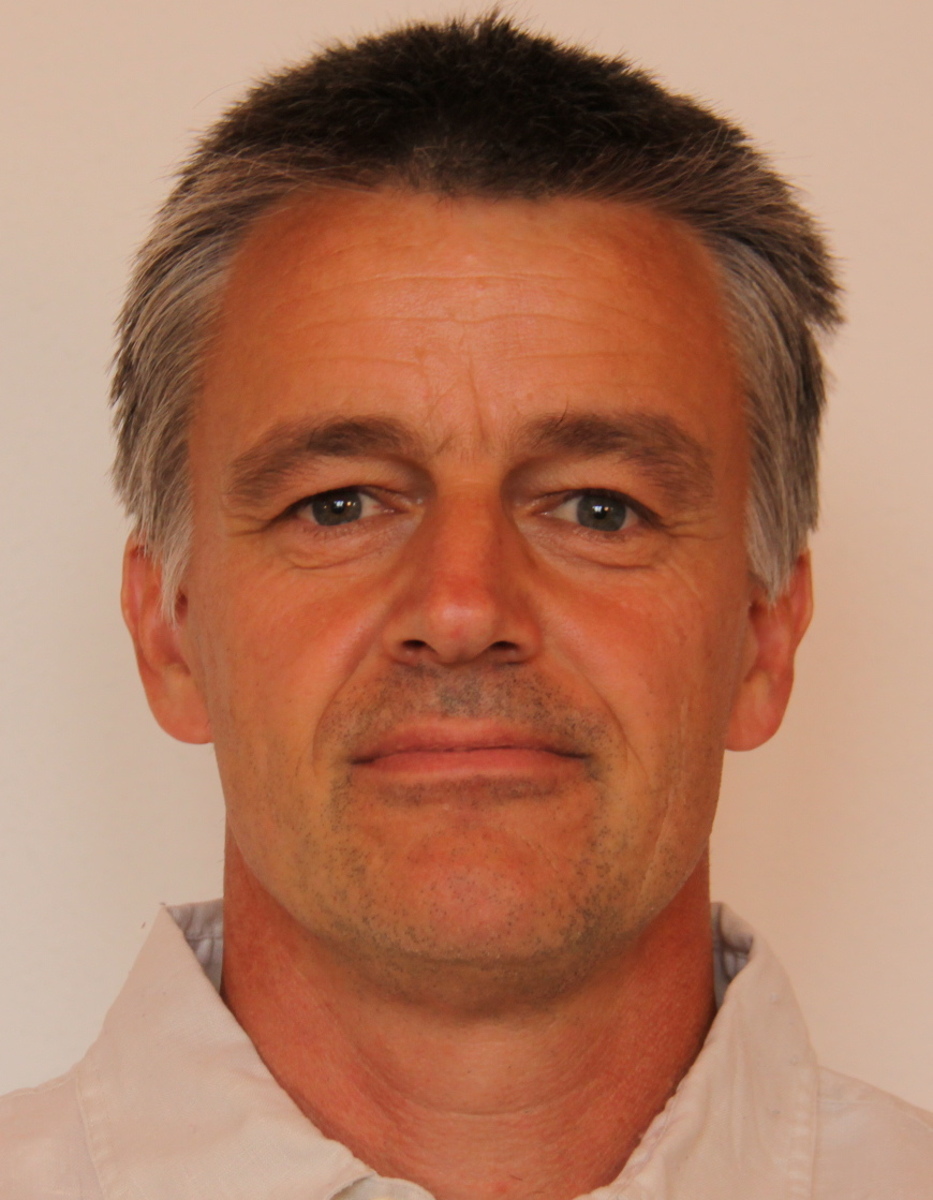}} \hfill
\subfloat[Morph]{\includegraphics[width=0.3\columnwidth]{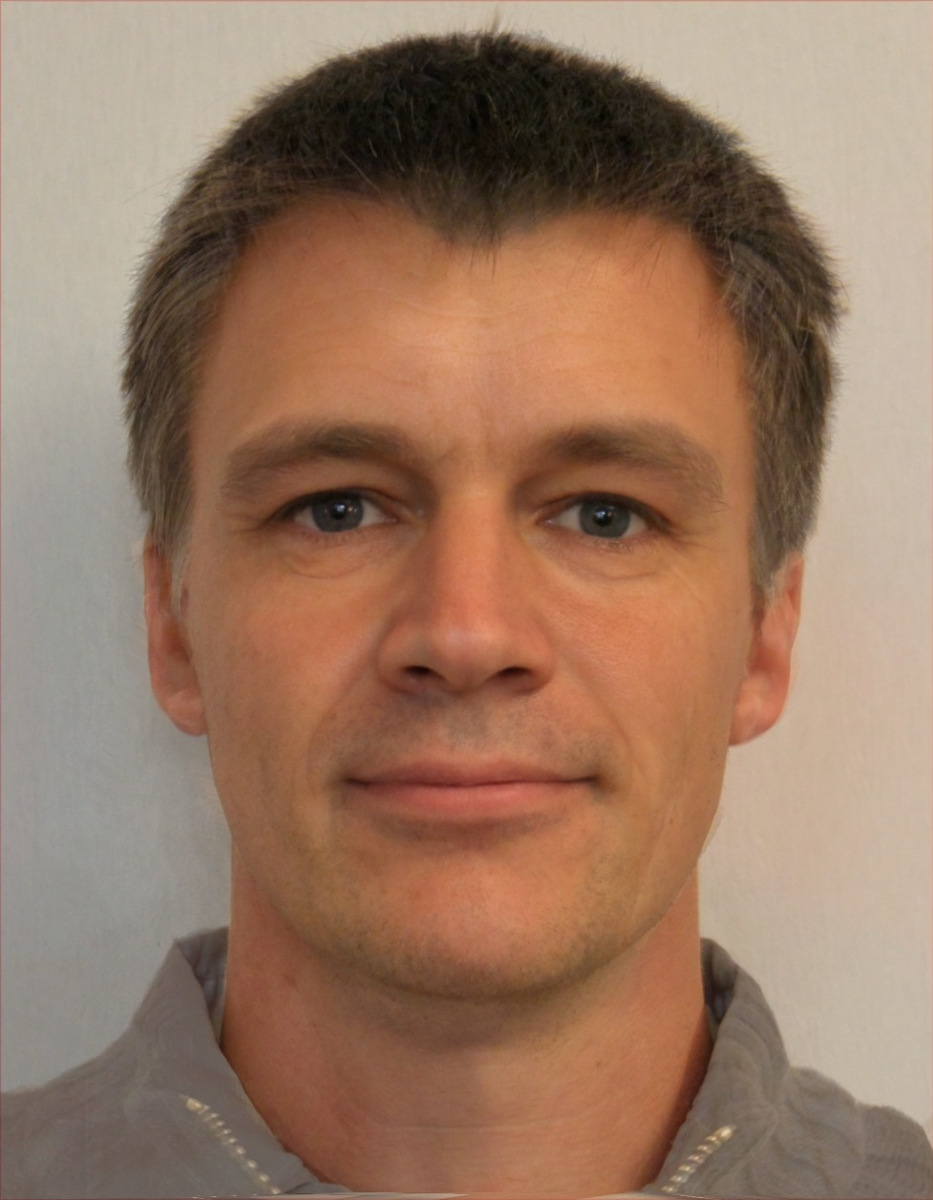}} \hfill
\subfloat[Subject 2]{\includegraphics[width=0.3\columnwidth]{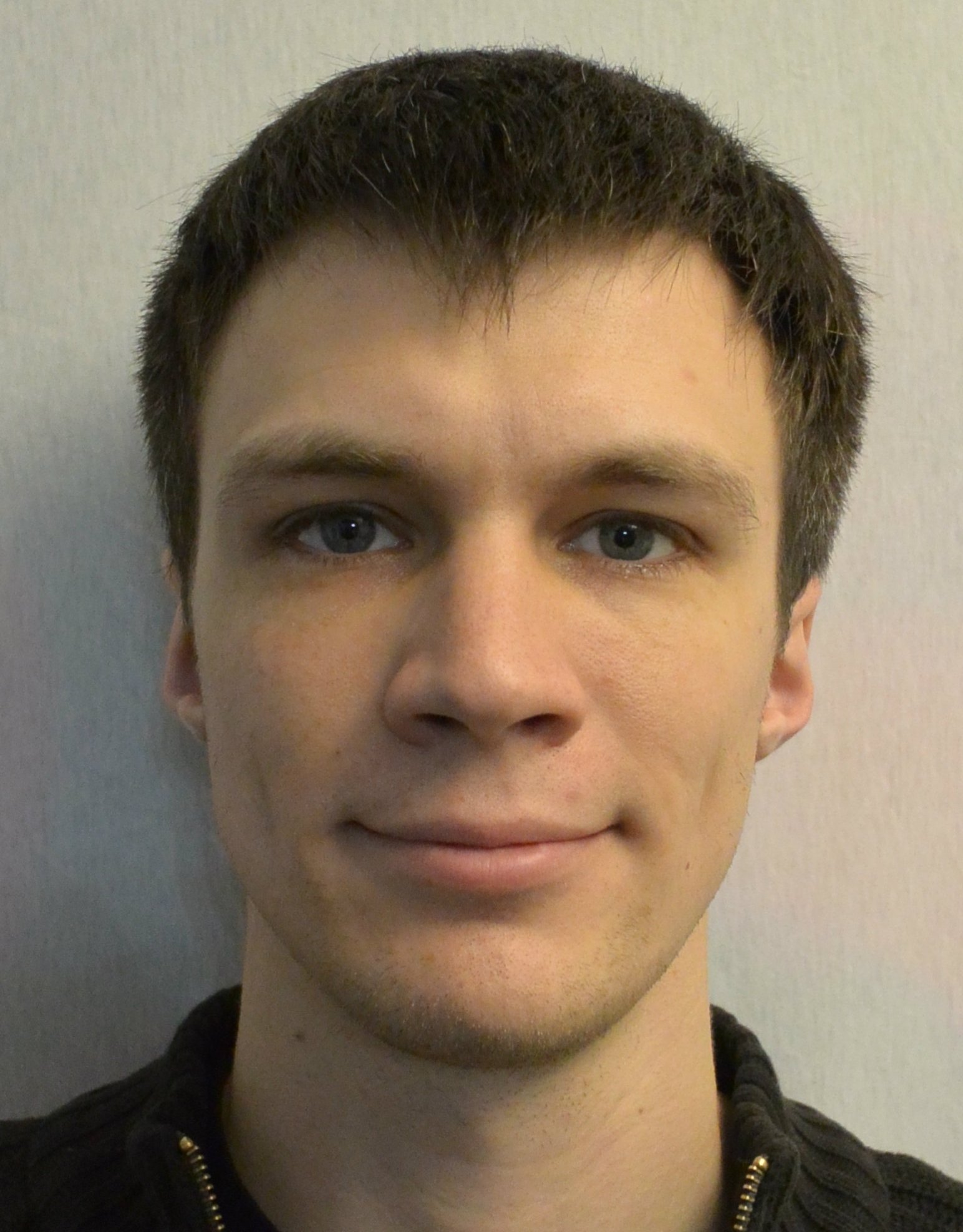}}
\caption{Morphing example (from Scherhag \etal \cite{Scherhag-MorphingVulnerability-2017})}
\label{fig:morph_example}
\end{figure}

\begin{figure*}[!ht]
\centering
\includegraphics[width=\textwidth]{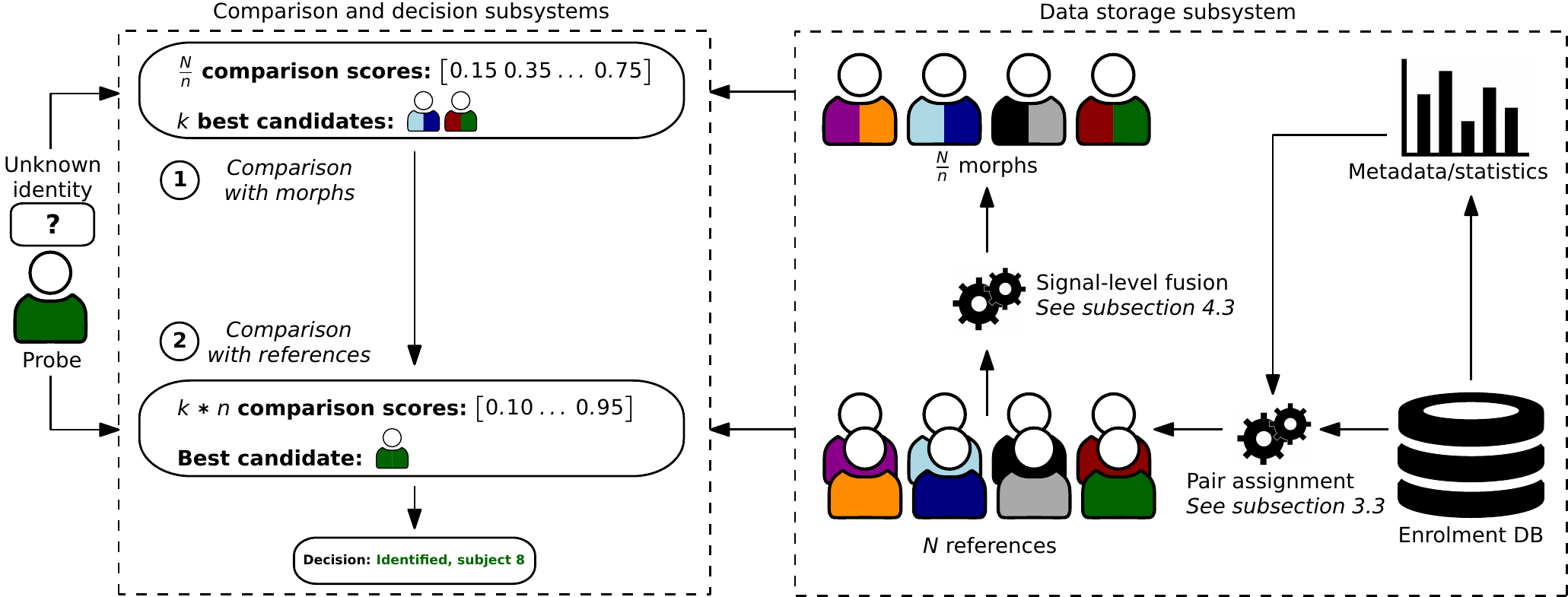}
\caption{Overview of the proposed system}
\label{fig:overview}
\end{figure*}

\section{Background and Related Work}
\label{sec:background}
The scope of this article combines two areas within biometrics, namely computational workload reduction and signal-level fusion through morphing. The pertinent background information and key related works for those two fields of research are surveyed in subsections \ref{subsec:background_workload} and \ref{subsec:background_morphing}, respectively.

\subsection{Computational Workload Reduction}
\label{subsec:background_workload}
To facilitate a growing number of data subjects (\ie the increase of the size of an enrolment database), optimisations or additional investment are needed to maintain fast biometric identification system response times. While expanding the hardware (\eg by distributing the computations to many servers) can solve the problem, such solutions are naturally associated with monetary costs for the equipment purchase, installation, maintenance, \etc In this context, the goal of computational workload reduction methods is to decrease the necessary amount of computations for a given task of a biometric system, thereby mitigating some of the physical infrastructure costs. Performing the biometric template comparisons dominates the overall computational costs of a biometric identification transaction; therefore, most of the workload reduction approaches address this step of the system pipeline \cite{Drozdowski-WorkloadSurvey-IET-2019}. Two main classes of such approaches are \textit{feature transformation}, with the aim to reduce the computational cost of individual template comparisons (see \eg \cite{Drozdowski-DeepFaceBinarisation-ICIP-2018}), and \textit{pre-selection}, with the aim of search space reduction, \ie the number of necessary template comparisons (see \eg \cite{Kavati-SearchReductionSurvey-IGI-2018}). In the context of this work, the latter type of methods is of interest.

A simple and common method for database pre-filtering is the utilisation of geographic and/or demographic metadata to narrow down the potential search space (see \eg Gehrmann \etal \cite{Gehrmann-MetadataFiltering-2019}); soft biometrics (see \eg Dantcheva \etal \cite{Dantcheva-SoftBiometricsSurvey-TIFS-2016}) can also be used in an analogous manner. The works of \cite{Gentile-TwoStageIris-BTAS-2009, Billeb-SpeakerTwoStage-BIOSIG-2014, Pflug-HistogramBinarisation-CYBCONF-2015} rely on a concept for a two-stage retrieval. In the first step, a compact representation (\eg binarised or with reduced dimensionality) of biometric data is used to pre-filter an enrolment database, whereas in the second step the actual (with high discriminative power) biometric templates are compared within the pre-selected shortlist. Similar methods based on coarse-to-fine search, nearest-neighbour search, and clustering based on the feature sets extracted from biometric samples have also been published. In addition to such single-modal methods, a number of methods which rely on biometric information fusion (see \eg \cite{Dinca-FusionSurvey-2017,Sing-Fusion-2019} for surveys on this topic) has been proposed. In \cite{Drozdowski-MultiFingerBinning-WIFS-2018} and \cite{Drozdowski-MultiIrisIndexing-IJCB-2017}, multi-instance binning and indexing methods were proposed for fingerprint and iris data, respectively. More generic multi-biometric indexing methods were proposed \eg in \cite{Jayaraman-MultimodalIndexing-ICISS-2008,Gyaourova-IndexCodes-CVPRW-2009,Gyaourova-IndexCodes-TIFS-2012}. Those methods relied on information fusion on the feature-level in order to build some kind of an intelligent search structure (\eg a tree). In \cite{Iloanusi-FingerprintIndexing-PRL-2014}, a multi-instance fingerprint indexing method which relies on rank-level fusion was proposed. A multi-biometric cascade where successively smaller candidate short-lists are created at score-level was proposed in \cite{Drozdowski-Kstage-2019}. A decision-based cascade operating on the principle of sequential fusion of fingerprint and iris recognition systems was presented in \cite{Elhoseny-CascadingFusion-2017}.

Notably absent in the aforementioned works (and indeed in a comprehensive survey of the field \cite{Drozdowski-WorkloadSurvey-IET-2019}) are approaches which rely on signal-level fusion. In the context of face recognition, signal-level fusion might be achieved with morphing (see subsection \ref{subsec:background_morphing}). While most of the contemporary research concentrates on the vulnerability caused by morphs (\ie average facial images derived from two or more parent images) and detection thereof, in this article they are considered from an entirely different angle -- their properties are exploited to reduce the computational costs of the biometric identification transactions. A short proof-of-concept for this idea was presented by Drozdowski \etal \cite{Drozdowski-MorphingPreselection-ICASSP-2019}, who coupled signal-level fusion (morphing) with a concept for a two-stage pre-filtering and retrieval system. This article is based on and significantly extends said work both conceptually and experimentally.

\subsection{Morphing}
\label{subsec:background_morphing}
Image morphing is a long-standing field of research with numerous practical applications, most notably in \eg medical imaging and special effects in the film industry \cite{Wolberg-ImageMorphing-1998}. In the context of biometrics, image morphing methods enable creation of biometric samples which \textit{contain biometric information from two or more distinct data subjects}. Such artificially created samples bear resemblance to the two (or more) original parent samples both in the feature and image domain. In other words, the unique link between individuals and their biometric reference data can be broken, as the subjects whose biometric samples are contained in the morphed image can both be matched (accepted) during subsequent biometric recognition transactions with the morphed reference image. Ferrara \etal \cite{Ferrara-MagicPassport-2014} showed that administrative processes for issuance of biometric travel documents have vulnerabilities (in certain countries), which enable a submission of a morphed image within a passport application process. After the (otherwise fully genuine) passport is produced, it makes it possible for multiple individuals to cross borders with biometric checks. This attack vector, dubbed the ``magic passport'', has been shown to be feasible both against automated systems and human experts alike \cite{Ferrara-FaceAlterations-2016}. Since then, it has also been shown that other biometric characteristics might be susceptible to attacks which rely on variations of morphing, see \eg \cite{Ferrara-FingerprintMorphing-2016,Rathgeb-IrisMorphing-IJCB-2017}.

The facial image morphing process is relatively simple; viable and realistic morphs can be generated even by non-experts with a variety of inexpensive or free software tools \cite{Scherhag-MorphingSurvey-2019}. A well-known case with practical relevance is that of a German activist group, ``Peng!'', who successfully applied for and received a passport with a morphed image of a group member and a high-ranking EU-level official \cite{Christie-GermanMorph-Vice-2018}. A typical morphing process consists of following steps: 1) facial landmark detection and triangulation in multiple input images, 2) landmark averaging to a single set of landmarks, 3) warping and alpha blending the information into a single output image, and 4) post-processing, such as image compression or artefact removal. Recently, morphing techniques based on generative adversarial networks (GANs) have been proposed, see \eg Damer \etal \cite{Damer-MorGAN-2018}. Figure \ref{fig:morph_example} shows an example of facial image morphing. 

In recent years, facial image morphing has been a hot topic in the biometric research community. Significant efforts were put into development of algorithms which can automatically and reliably detect morphed images. Survey articles by Makrushin \etal \cite{Makrushin-FaceMorphing-2018} and Scherhag \etal \cite{Scherhag-MorphingSurvey-2019} provide a detailed overview of facial morph creation and detection methods. Those are, however, out of scope for this work; instead, the intention of this work is to take advantage of morphing with the goal of improving a biometric system. Preliminary works on this subject included \eg Korshunov \etal \cite{Korshunov-MorphingPrivacy-2013}, who used morphing for privacy-preservation, and Drozdowski \etal \cite{Drozdowski-MorphingPreselection-ICASSP-2019}, who conducted a proof-of-concept study on accelerating biometric identification transactions using morphing.

\section{Proposed System}
\label{sec:proposedsystem}
In this section, the proposed system is described. Subsection \ref{subsec:proposedsystem_overview} provides a conceptual overview of the proposed system. Subsections \ref{subsec:proposedsystem_retrieval} and \ref{subsec:proposedsystem_pairselection} describe the algorithms for retrieval in an identification transaction and selection of subjects to form morph pairs, respectively.

\subsection{Overview}
\label{subsec:proposedsystem_overview}
The proposed system relies on signal-level fusion (facial image morphing) to create a multi-stage retrieval structure for biometric identification transactions.  The requisite changes \wrt a simple, exhaustive search-based biometric identification system mainly pertain to two subsystems of a generic biometric system as specified in ISO/IEC 19795-1 \cite{ISO-PerformanceReporting-2021}:

\begin{LaTeXdescription}
\item[Data storage subsystem] contains the biometric samples representing $N$ enrolled subjects, from which an additional index is created through the application of signal-level fusion, whereby the resulting index-samples contain biometric information from multiple subjects (recall subsection \ref{subsec:background_morphing}). This signal-level fusion can be performed for two or multiple parent samples, so that each sample contributes equally to the resulting morph. The number of parent samples contributing to a morph is referred to as ``morph capacity'' and denoted as $n$; thus, $\frac{N}{n}$ morphs are created. In subsection \ref{subsec:proposedsystem_pairselection}, the methods for pairing the samples to be morphed are described and analysed.
\item[Comparison subsystem] the biometric probe is first compared against the $\frac{N}{n}$ morphed samples exhaustively. A short-list of the most likely $k$ candidates is pre-selected based on those comparison scores. On a secondary processing level, template comparisons between the biometric probe and the normal reference templates are conducted within the candidate short-list.
\end{LaTeXdescription}

Figure \ref{fig:overview} shows a conceptual overview of the proposed system. It is worth noting that the proposed system could be effortlessly combined with certain other methods of computational workload reduction, such as binning or pre-selection based on demographic or geographic attributes. More precisely, the combination with said methods could significantly reduce the computational effort required in the first step of the retrieval algorithm of the proposed system (see next subsection).

\subsection{Retrieval}
\label{subsec:proposedsystem_retrieval}
The proposed system relies on the fact that the morphed images retain enough discriminative power for the probe to exhibit better comparison scores against its correct (mated) morph than against the other (non-mated) morphs. This in turn makes it possible to robustly select a candidate short-list to be passed onto the next stage of the pipeline. The computational workload of an identification transaction (measured in terms of the number of necessary template comparisons and denoted $W_{\mathrm{two-stage}}$), can be expressed as follows:

\begin{equation}
\label{eq:comparisons_twostage}
W_{\mathrm{two-stage}} = \frac{N}{n} + k * n
\end{equation}

\begin{figure}[!ht]
\centering
\includegraphics[width=\columnwidth]{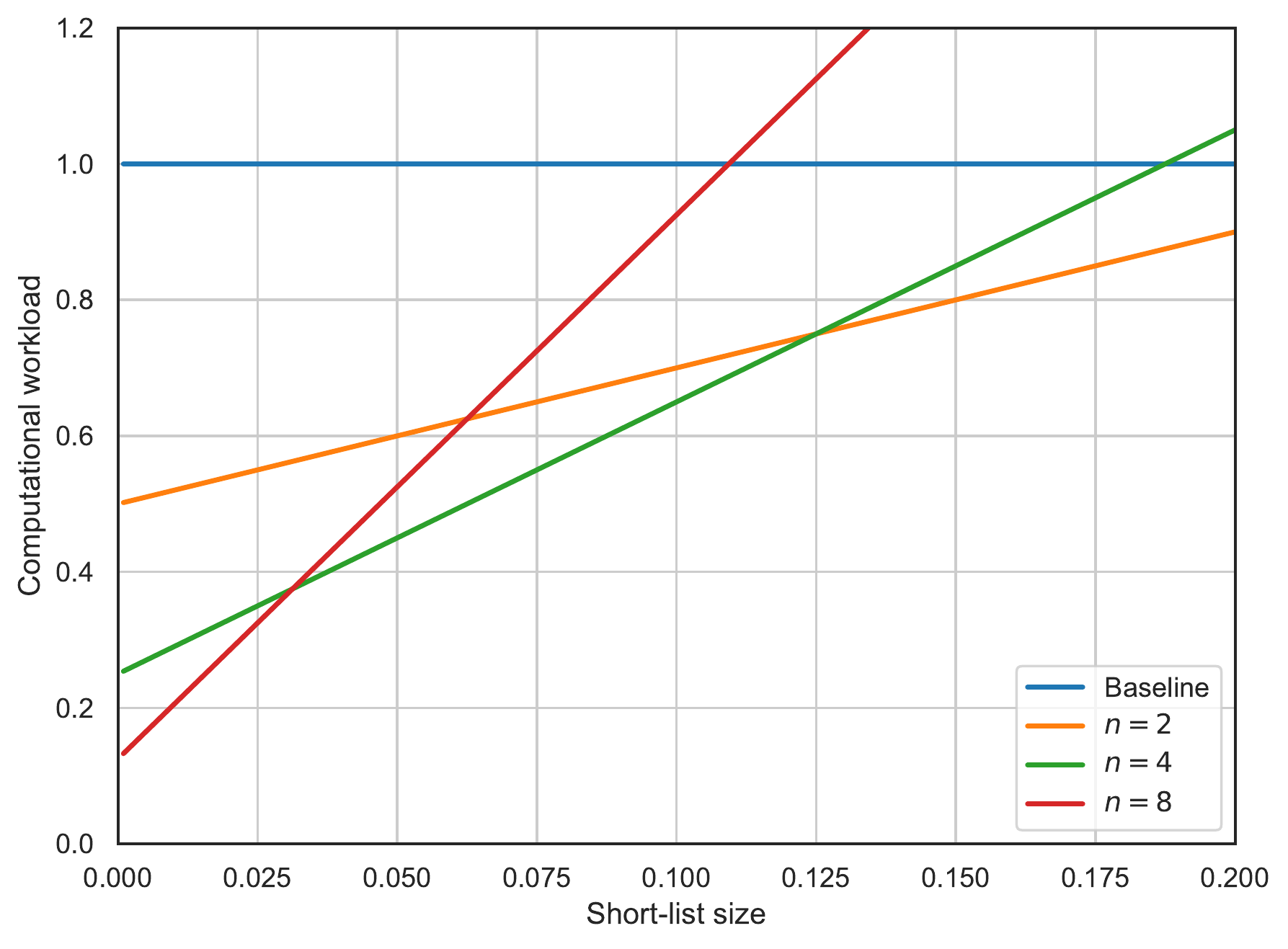}
\caption{Computational workload per identification transaction of the proposed two-stage (1 stage with morphed images and 1 stage with reference images) system in relation to an exhaustive search-based baseline system. The parameter $k$ is implicitly included in the calculation of x-axis values.}
\label{fig:comparisons_example_twostage}
\end{figure}

A decision space for the computational workload of the proposed system can be visualised by controlling the $n$ and $k$ parameters. This is shown in figure \ref{fig:comparisons_example_twostage}. The y-axis expresses a fraction of the baseline system's computational workload (as proposed in \cite{Drozdowski-BloomFilterIndexing-IET-2018}); the baseline workload is equal to $1.0$. The x-axis expresses the size of the pre-selected candidate short-list as a fraction of the enrolment database size (\ie $\frac{k}{N}$). The baseline (exhaustive search) does not depend on the aforementioned parameters and is therefore plotted as a horizontal line for reference. It can be seen that provided a sufficiently small short-list (\ie $\frac{k}{N} < 0.1$), the proposed system might significantly reduce the computational workload of a biometric identification transaction. 

In addition to a two-stage retrieval, the concept can be extended to facilitate multiple stages. In such a scheme, the candidate short-list is successively filtered in a cascading manner (conceptually similar to the multi-stage and multi-biometric cascade of Drozdowski \etal \cite{Drozdowski-Kstage-2019}), with the number of subjects contributing to the morphs being reduced at each level. Instead of a single threshold for the short-list size, each level would require its own threshold. Such a multi-stage system depends on following variables: $N$, the number of subjects represented in the enrolment database; $n_{1}$ selected from the set $\set{2^{x} \mid x \in \mathbb{N^{+}}}$ and denoting the number of subjects contributing to the morphs on the first level, thereby also determining the total number of levels of the cascade, \ie $l = \log_{2} n_{1} + 1$; and the short-list size thresholds for each cascade level, \ie $\set{k_{1} \dots k_{l}}$. The computational workload in a multi-stage retrieval scenario, denoted $W_{\textrm{multi-stage}}$, can be obtained using the following formula:

\begin{equation}
\label{eq:comparisons_multistage}
W_{\mathrm{multi-stage}} = \frac{N}{n_{1}} + \sum_{l=2}^{\log_{2} n_{1}} 2k_{l}
\end{equation}

\begin{figure}[!ht]
\centering
\includegraphics[width=\columnwidth]{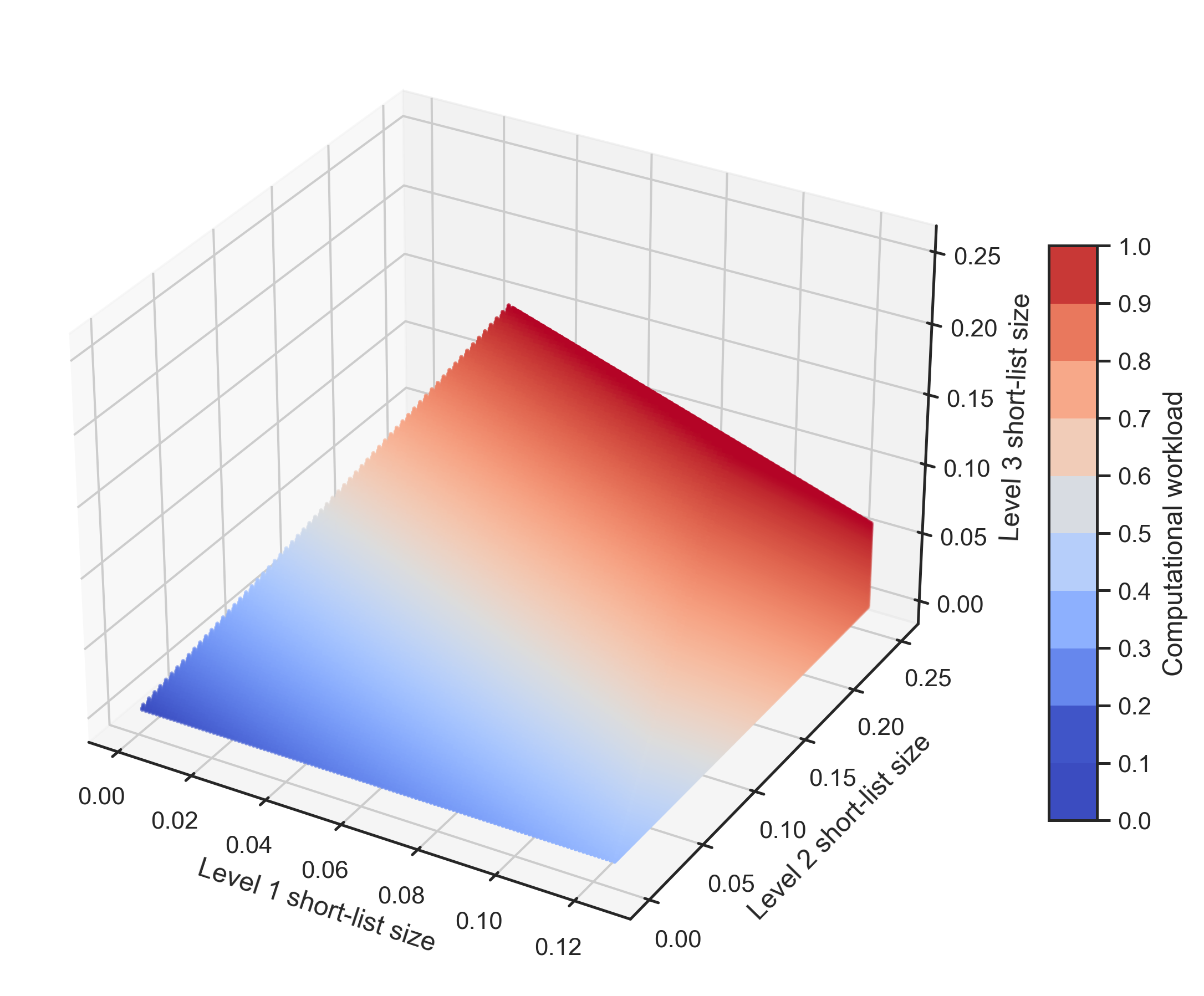}
\caption{Computational workload per identification transaction of the proposed multi-stage system (3 stages with morphed images and 1 stage with reference images) in relation to an exhaustive search-based baseline system. The parameters $k_{l}$ are implicitly included in the calculation axis values corresponding to their level ($l$) in the multi-stage system.}
\label{fig:comparisons_example_multistage}
\end{figure}

A decision space for the computational workload of the proposed multi-stage system for a cascade with 3 levels of morphs (\ie starting with morphs each consisting of 8 data subjects) and a final level with the actual reference images can be drawn as shown in figure \ref{fig:comparisons_example_multistage}. The x, y, and z-axes denote the short-list size (\ie the $k_{l}$ parameter in relation to $N$) on each level of the cascade, whereas the colour space denotes the overall workload in relation to the exhaustive search-based baseline, whose workload equals $1.0$. For clarity, only the configurations resulting in computational workload equal to or lower than the baseline are depicted. A cascade with 2 levels of morphs (\ie starting with morphs each consisting of 4 data subjects) is also theoretically feasible, whereas a cascade with more than 3 levels of morphs does not appear theoretically feasible. This limitation is due to the prohibitively lowered discriminative power of the system as too much information is lost by morphing so many (\ie 16 or more) subjects with each other.

In order to reduce the computational workload of an identification transaction, the proposed system must require fewer template comparisons than a typical, exhaustive search-based one. In other words, the relation $W_{\mathrm{proposed}} < N$ must be satisfied. From the theoretical calculations demonstrated in figures \ref{fig:comparisons_example_twostage} and \ref{fig:comparisons_example_multistage} for the proposed two-stage and multi-stage systems, respectively, it can be seen that there do exist configurations (based on $k$ and $n$ parameters) which require significantly less computational workload than the baseline. In other words, provided that the discriminative power (\ie biometric performance) of the system can be maintained, the proposed system could operate more computationally efficiently than the baseline. This trade-off between computational workload and biometric performance is evaluated empirically later on in this article.

\subsection{Selection of Morph Pairs}
\label{subsec:proposedsystem_pairselection}
Deciding which parent samples to morph with each other is expected to have a non-trivial impact on the efficacy of the proposed system. This assumption is based on previous works by Damer \etal \cite{Damer-MorphingPairs-2019} and R{\"o}ttcher \etal \cite{Rottcher-MorphingDoppelganger-2020}, who showed that morphing attacks can be improved by intelligently (\ie based on quantitative criteria) selecting subjects to morph with each other. 

Such a subject-pair selection belongs to a well-known and old class of combinatorial optimisation problems. It could be, for example, formulated as a stable roommates/marriage problem. However, in the practical experiments, this formulation ran into a number of issues regarding so-called ``odd pairs'' and solvability of the problem for a large number of data subjects (see \cite{Pittel-StableRoommates-1994,Chung-StableRoommates-2000}). To circumvent this issue, instead of seeking a stable matching, one could try finding a matching based on a global (\ie for the entire enrolment database) optimisation of a cost function, thereby allowing some poorly matched pairs (\ie with a high cost). This formulation corresponds to the assignment problem and was successfully applied in the practical experiments.

Formally, given $S$ as the set of data subjects in the enrolment database and a weight function $C: S \times S \to \mathbb{R}^{+}$, the goal is to find a bijective mapping of this set to itself, $f: S \to S$, so that the cost function $\displaystyle\sum_{s \in S} C_{s, f(s)}$ is minimised and $\forall s \in S, f(s) \neq s$ (\ie the subjects cannot be mapped to themselves). In this work, three selection methods were considered:

\begin{LaTeXdescription}
\item[Random] the pairs of samples to be morphed are assigned by chance, \ie without any kind of selection.
\item[Soft-biometric] the pairs of samples to be morphed are assigned based on similarity in terms of soft-biometric attributes. In this work, the subject's sex, age, and skin colour were used.
\item[Similarity-score] the pairs of samples to be morphed are selected based on similarity in terms of comparison scores computed with a facial recognition system.
\end{LaTeXdescription}

For the latter two, the weight function is obtained by calculating the dissimilarity (soft-biometric or face recognition-based) scores. Thus, for an enrolment database of $N$ subjects, a square matrix is created as shown in equation \ref{eq:scores_matrix}, where $S_{x}$ denotes the $x$'th data subject, while $c_{x,y}$ denotes the dissimilarity score between the $x$'th and $y$'th data subject in the enrolment database (\ie the cost of pairing the two subjects with each other). Due to the constraint that the subjects cannot be paired with themselves, the diagonal is set to $\infty$ (represented by the floating-point format size limit in the actual software implementation).

\begin{equation}
\label{eq:scores_matrix}
C = \bordermatrix{~ & S_{1} & S_{2} & S_{3} & \cdots & S_{N} \cr
                  S_{1} & \infty & c_{1,2} & c_{1,3} & \cdots & c_{1,N} \cr
                  S_{2} & c_{2,1} & \infty & c_{2,3} & \cdots & c_{2,N} \cr
                  S_{3} & c_{3,1} & c_{3,2} & \infty & \cdots & c_{3,N} \cr
                  \vdots & \vdots & \vdots & \vdots & \ddots & \vdots \cr
                  S_{N} & c_{N,1} & c_{N,2} & c_{N,3} & \cdots & \infty \cr
                  }
\end{equation}

Using the above matrix-based formulation, this problem can be solved in polynomial time with the so-called Hungarian algorithm \cite{Kuhn-HungarianAlgorithm-1955}. The expectation of such an intelligent assignment is an increased discriminative power of the first step(s) (\ie the pre-selection) of the proposed system, whereby its overall results (biometric performance and computational workload) would be improved. For configurations where more than two data subjects contribute to the morphs in the cascade, \ie $n > 2$, the pairing follows an iterative procedure. First, the pairings are computed for individual reference images as described above and two-subject morphs are created accordingly. Thereupon, the procedure is repeated for the resulting morphs (\ie the cost matrix is re-computed and the morphing is done) in order to create morphs with four or eight contributing subjects.

\section{Experimental Setup}
\label{sec:experimentalsetup}
The following subsections describe the experimental setup used for the evaluation of the proposed system. The used datasets are presented in subsection \ref{subsec:experimentalsetup_datasets}, while the used face recognition systems and image morphing tools are presented in subsections \ref{subsec:experimentalsetup_facerecognitionsystems} and \ref{subsec:experimentalsetup_morphingtools}, respectively. Finally, subsection \ref{subsec:experimentalsetup_evaluationmetrics} specifies the used evaluation metrics.

\begin{table}[!ht]
\centering
\caption{Summary of experimental parameter counts}
\label{table:combinations}
\resizebox{\columnwidth}{!}{
\begin{tabular}{ll}
\toprule
\textbf{Parameter} & \textbf{Count} \\ 
\midrule
Number of subjects in a morphed image (subsection \ref{subsec:proposedsystem_retrieval}) & 3 \\ 
Image pairing method for morphing (subsection \ref{subsec:proposedsystem_pairselection}) & 3 \\
Face recognition system (subsection \ref{subsec:experimentalsetup_facerecognitionsystems}) & 4 \\
Image morphing tool (subsection \ref{subsec:experimentalsetup_morphingtools}) & 4 \\
\bottomrule
\end{tabular}
}
\end{table}

All possible combinations of the face recognition and morphing tools, as well as settings of the proposed system were evaluated. For the random pairing algorithm, a cross-validation over 10 folds was conducted. Table \ref{table:combinations} summarises the counts of the individual parameters; accordingly, the total number of tested configurations was 144.

\subsection{Datasets}
\label{subsec:experimentalsetup_datasets}
\begin{figure}[!ht]
\centering
\subfloat[FERET\label{fig:example_feret}]
{
\includegraphics[width=0.235\columnwidth]{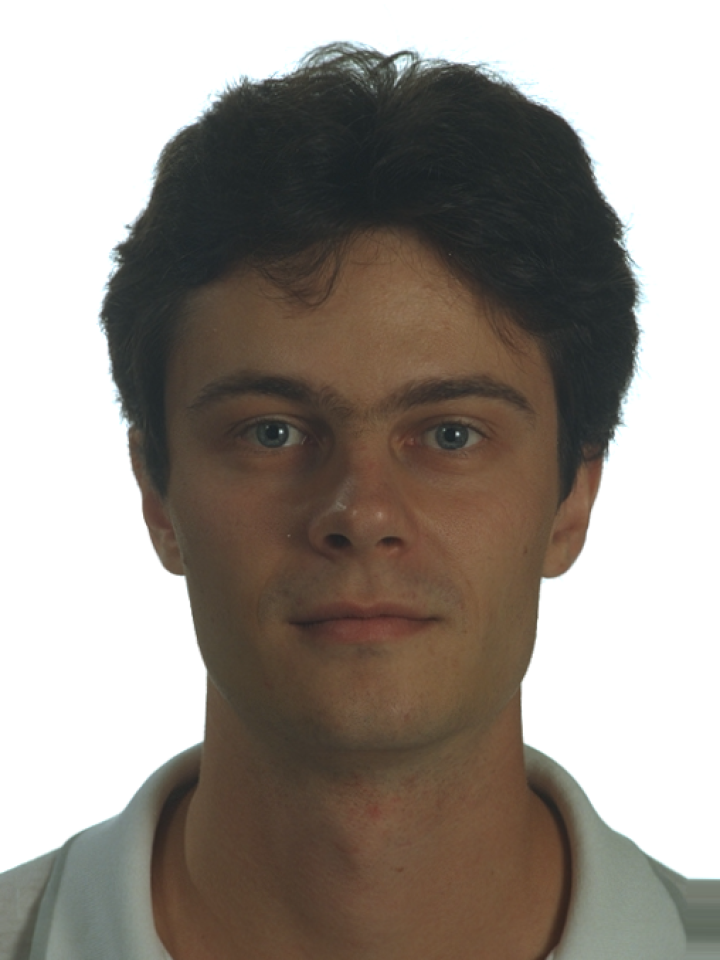} \hfill
\includegraphics[width=0.235\columnwidth]{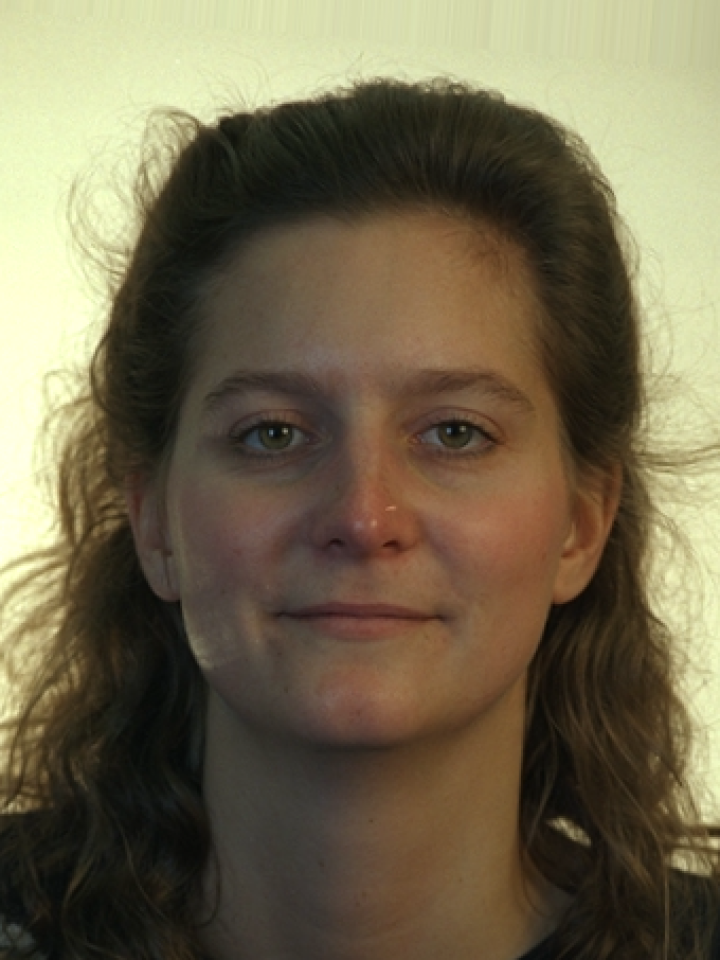} \hfill
\includegraphics[width=0.235\columnwidth]{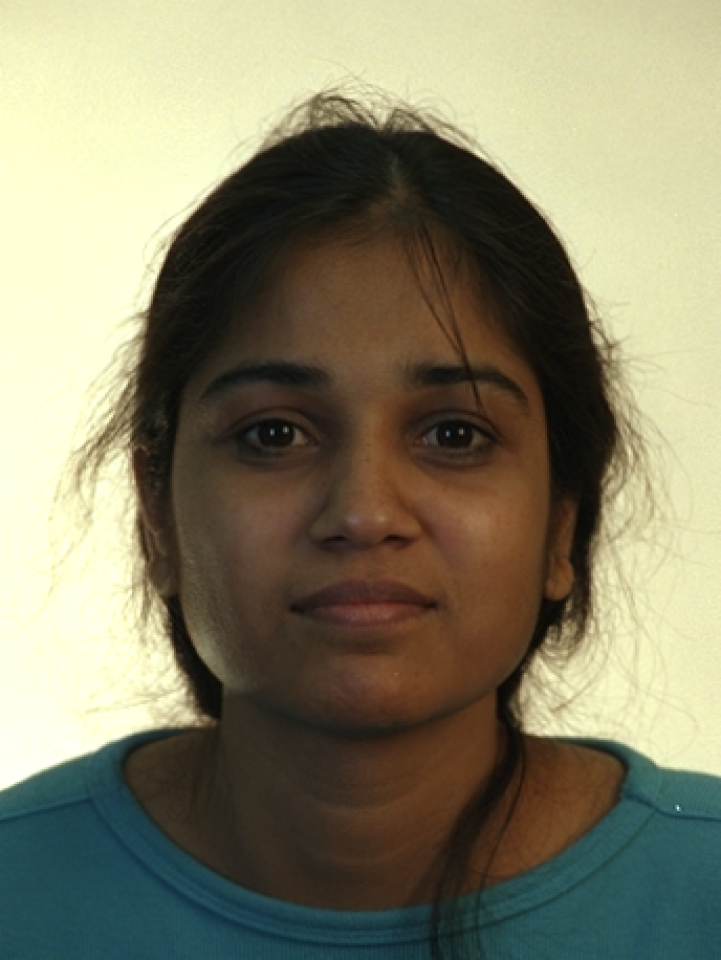} \hfill
\includegraphics[width=0.235\columnwidth]{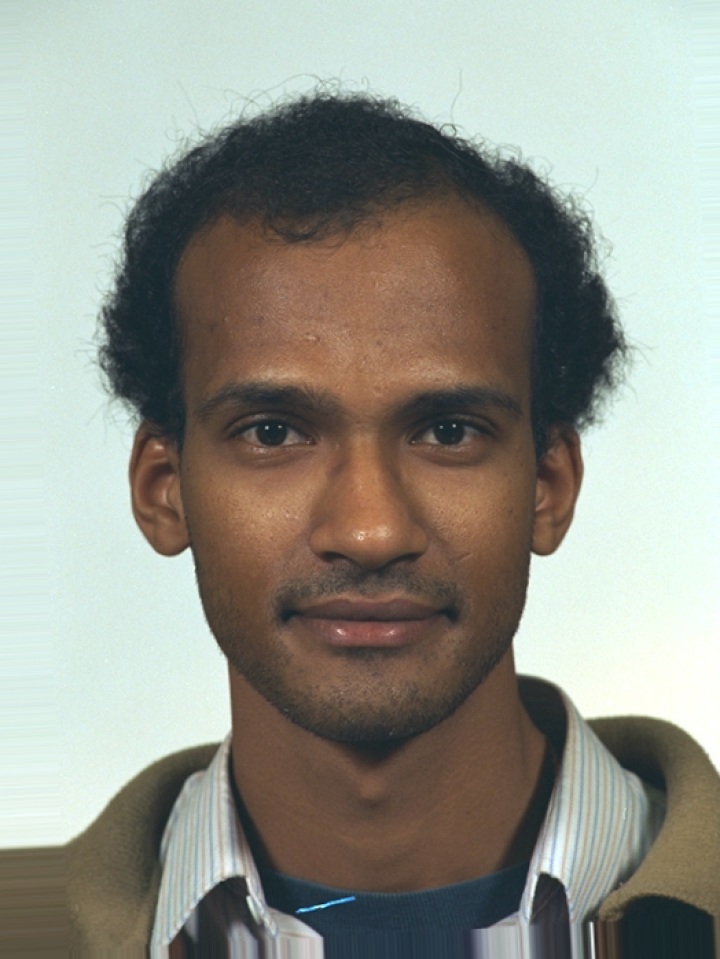}
} \\
\subfloat[FRGCv2\label{fig:example_frgc}]
{
\includegraphics[width=0.235\columnwidth]{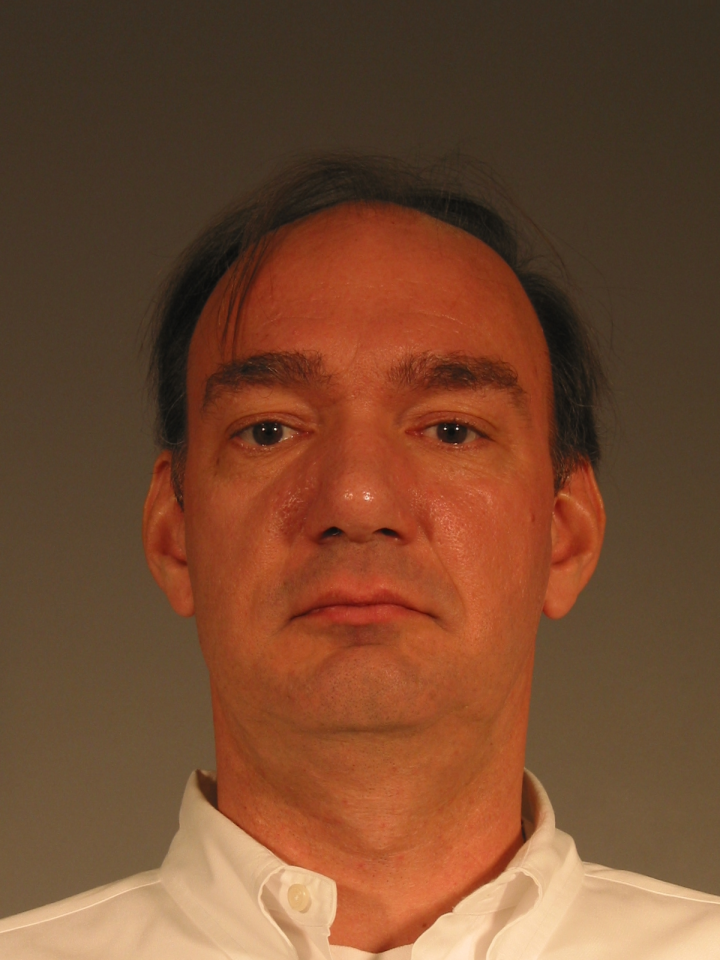} \hfill
\includegraphics[width=0.235\columnwidth]{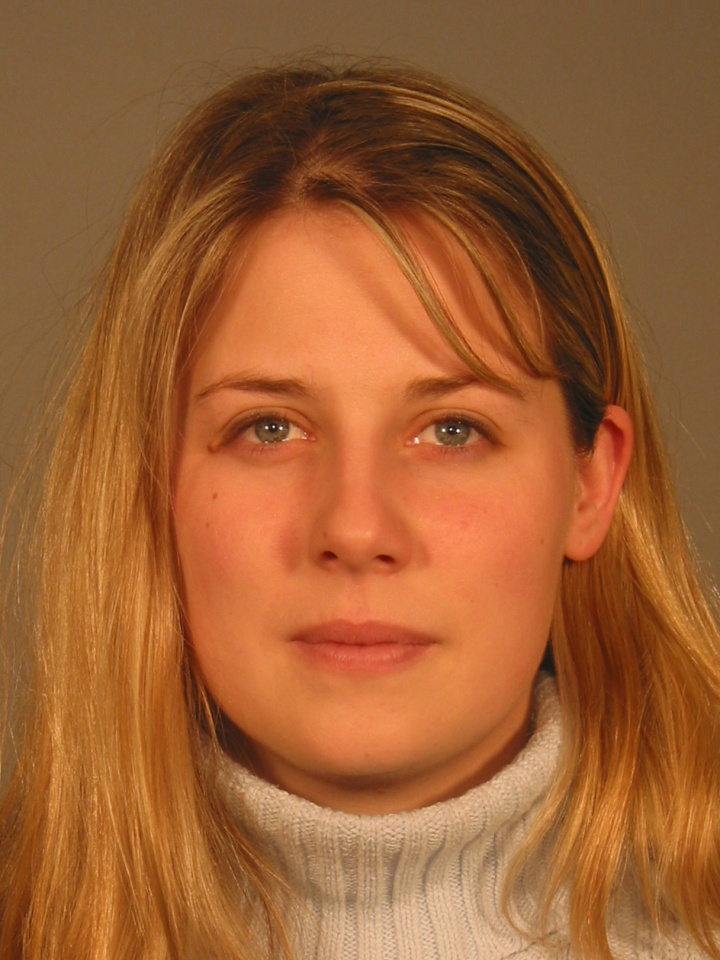} \hfill
\includegraphics[width=0.235\columnwidth]{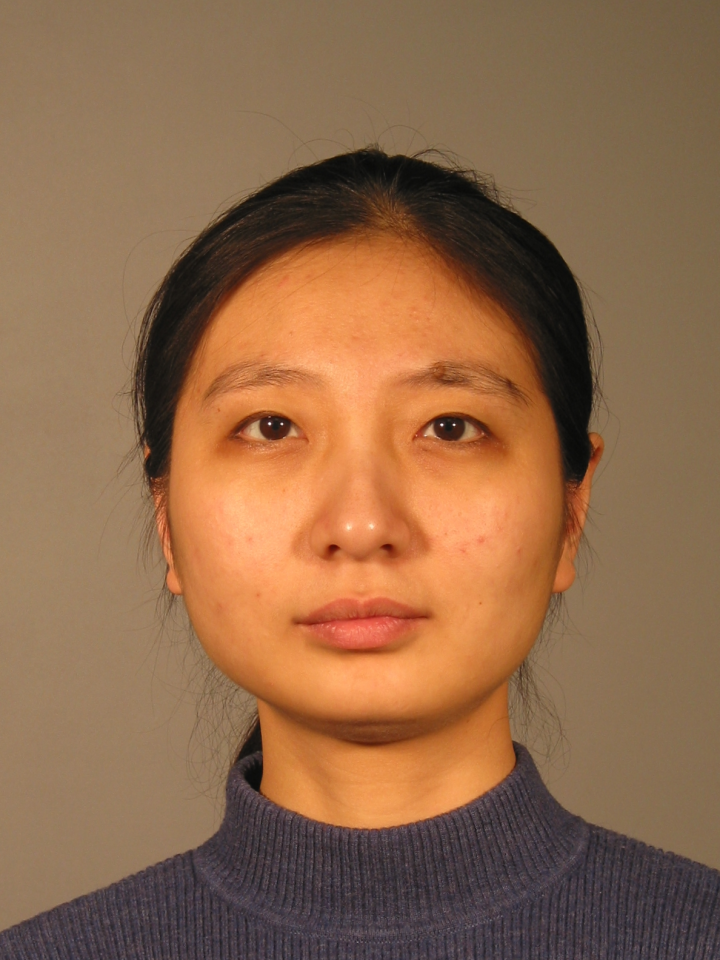} \hfill
\includegraphics[width=0.235\columnwidth]{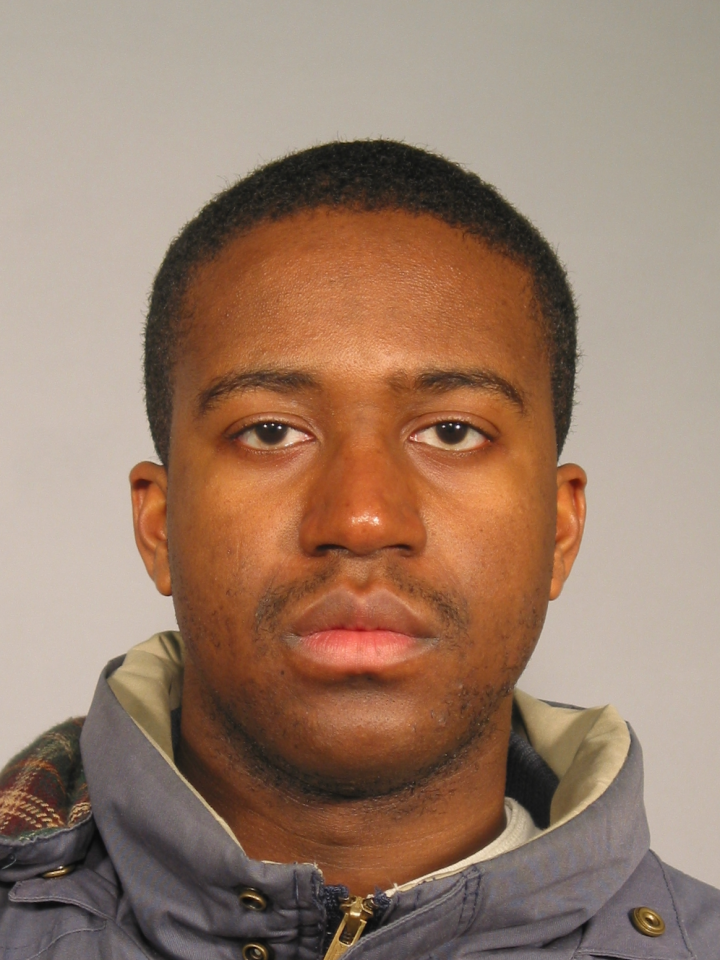}
}
\caption{Example images from the used datasets}
\label{fig:example_images}
\end{figure}

The experiments were conducted using a combined database of over 1,000 subjects stemming from the FERET \cite{Phillips-FERET-TPAMI-2000} and FRGCv2 \cite{Phillips-FRGC-CVPR-2005} databases. A subset of images was selected based on conformance with ICAO requirements for passport images \cite{ICAO-9303-p9-2015}. To facilitate the soft-biometric pairing method described in subsection \ref{subsec:proposedsystem_pairselection}, the datasets were annotated for the following demographic attributes (where available, existing labels given by dataset providers were used directly): sex, age, skin colour. Table \ref{table:datasets} gives an overview of the used datasets, while figure \ref{fig:example_images} shows example images from them.

\begin{table}[!ht]
\centering
\caption{Overview of the used datasets' subsets}
\label{table:datasets}
\resizebox{0.5\columnwidth}{!}{
\begin{tabular}{lll}
\toprule
\textbf{Name} & \textbf{Subjects} & \textbf{Images} \\ 
\midrule
FERET & 530 & 1,414 \\
FRGCv2 & 532 & 3,109 \\
\bottomrule
\end{tabular}
}
\end{table}

It should be noted that since the morphing process requires highly constrained images, using in-the-wild datasets is not applicable in the context of the conducted research.

\subsection{Face Recognition Systems}
\label{subsec:experimentalsetup_facerecognitionsystems}
Several well-known and/or state-of-the-art deep learning-based open-source face recognition systems were used. Specifically, FaceNet \cite{Schroff-Facenet-CVPR-2015}, VGGFace2 \cite{Parkhi-VGGFace-BMVC-2015}, and ArcFace \cite{Deng-ArcFace-2018} with pre-trained models made available by the authors were utilised. Additionally, a very recent version of a strong commercial off-the-shelf (COTS) system from a well-established provider was utilised. For the used open-source systems, access to the underlying algorithms and feature representations is available, whereas the COTS system operates in a black-box manner and only allows the comparison score between two biometric samples to be computed.

\subsection{Morphing Tools}
\label{subsec:experimentalsetup_morphingtools}
\begin{figure}[!ht]
\centering
\subfloat[Subject pair 1]{
\includegraphics[width=0.225\columnwidth]{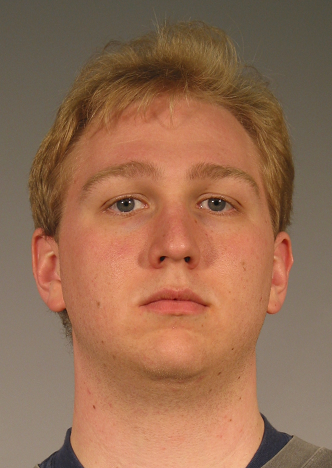}
\includegraphics[width=0.225\columnwidth]{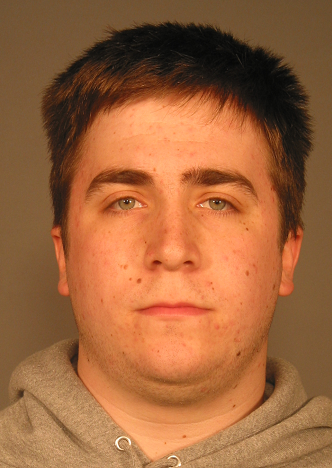}
} \hfill
\subfloat[Subject pair 2]{
\includegraphics[width=0.225\columnwidth]{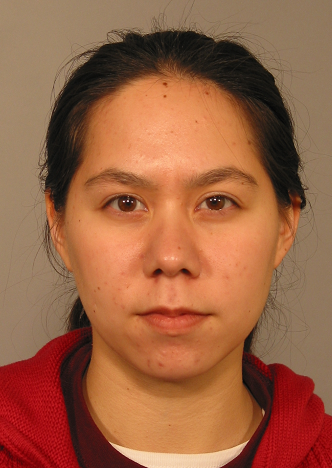}
\includegraphics[width=0.2325\columnwidth]{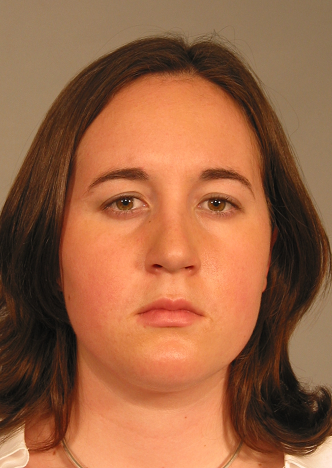}
} \\
\subfloat[OpenCV]{
\includegraphics[width=0.225\columnwidth]{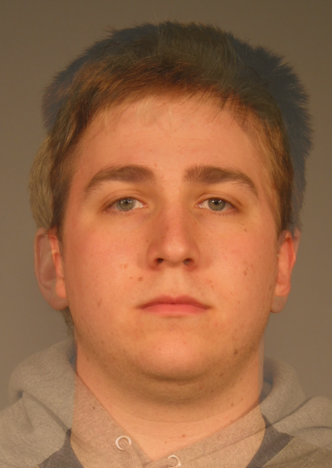}
\includegraphics[width=0.225\columnwidth]{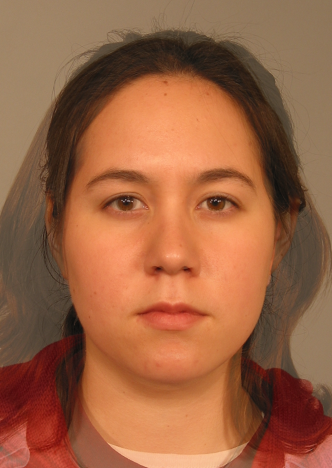}
} \hfill
\subfloat[FaceFusion]{
\includegraphics[width=0.225\columnwidth]{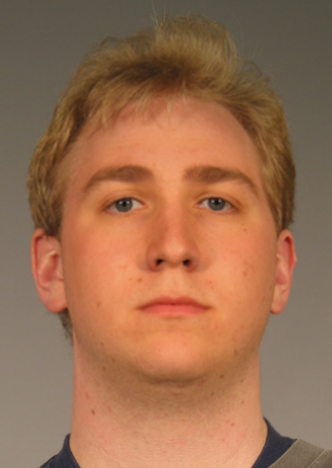}
\includegraphics[width=0.225\columnwidth]{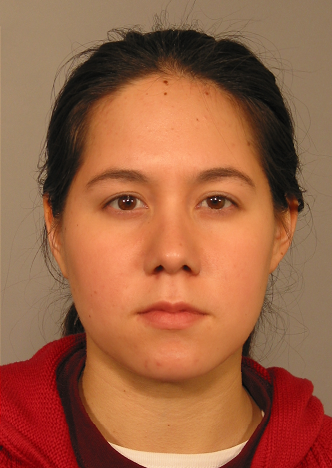}
} \\
\subfloat[FaceMorpher]{
\includegraphics[width=0.225\columnwidth]{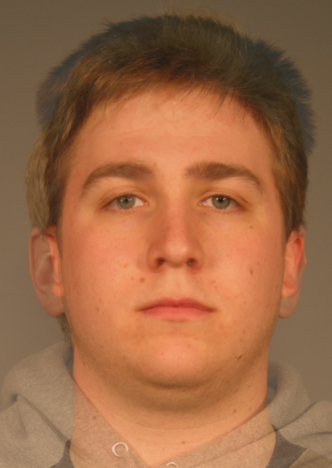}
\includegraphics[width=0.225\columnwidth]{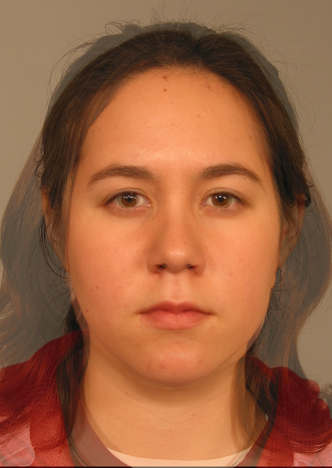}
} \hfill
\subfloat[UBO-Morpher]{
\includegraphics[width=0.225\columnwidth]{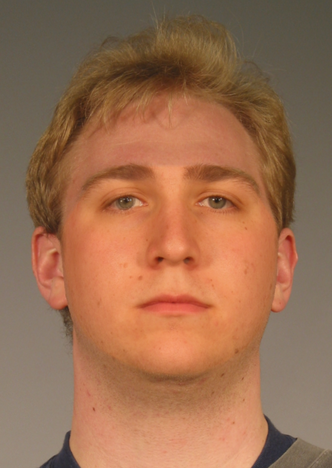}
\includegraphics[width=0.225\columnwidth]{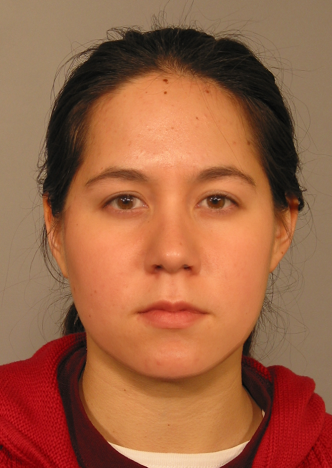}
}
\caption{Example morphs produced by the used tools (images in subfigures (a) and (b) taken from FRGCv2)}
\label{fig:example_morphs}
\end{figure}

Four tools were used to create morphed images for the experiments: the open-source OpenCV \cite{Morphing-Opencv-2016} and FaceMorpher \cite{Alyssa-Facemorpher} implemented with Python and dlib \cite{King-dlib-2009}, as well as the proprietary FaceFusion \cite{MomentMedia-Facefusion} and UBO-Morpher \cite{UBO-Morpher}. The resulting images exhibit differences due to pre- and post-processing being applied (\eg filters, cropping) and variations in the underlying morphing algorithms. Example images of morphs produced by the used tools are shown in figure \ref{fig:example_morphs}. For the pair selection method described in subsection \ref{subsec:proposedsystem_pairselection}, a SciPy \cite{Virtanen-SciPy-2020} implementation of the Hungarian algorithm was used.

\subsection{Evaluation Metrics}
\label{subsec:experimentalsetup_evaluationmetrics}
The proposed method and an exhaustive-search based baseline method were evaluated on two key aspects, using ISO/IEC standard methods and metrics \cite{ISO-PerformanceReporting-2021} as well as additional, commonly used ones:

\begin{LaTeXdescription}
\item[Biometric performance] For closed-set experiments, the hit rate (HR) and rank-1 recognition rate (RR-1); for open-set experiments, the decidability index $d'$ \cite{Daugman-DecisionLandscapes-UCAM-2000}, the equal-error-rate (EER) and false negative identification rate at a certain (here 0.1\%) false positive identification rate (denoted $\text{FNIR}_{0.1\%}$). 
\item[Computational workload] the overall computational workload (W) of a biometric identification transaction as a percentage of the baseline (exhaustive search) workload in terms of the number of template comparisons.
\end{LaTeXdescription}

\section{Results}
\label{sec:results}
The following subsections present the results of the conducted experimental evaluations. First, in subsections \ref{subsec:results_selectingrecognition} and \ref{subsec:results_selectingmorphing}, experiments are conducted \wrt the choice of suitable face recognition and image morphing methods. The effects of subject pairing methods are evaluated in subsection \ref{subsec:results_impactpairing}. Finally, the results achieved by the proposed pre-selection and overall multi-step system are presented in subsection \ref{subsec:results_overall}.

\subsection{Selecting Face Recognition Systems}
\label{subsec:results_selectingrecognition}
The systems described in subsection \ref{subsec:experimentalsetup_facerecognitionsystems} were first evaluated in a baseline scenario, where closed and open-set identification was conducted on an enrolment database of 1024 data subjects using exhaustive search method (hence $\text{W}_{\text{Baseline}} = 100\%$). Those results are reported in table \ref{table:results_baseline}. It can be seen that the COTS system and the open-source ArcFace system achieve high separability of the score distributions and perform well in both operation modes, whereas the biometric performance is strongly degraded for the other two open-source systems. This is especially the case for the open-set mode, which is much more challenging than closed-set. For the experimental results in the subsequent subsections, the COTS system is used along with the best open-source software system (ArcFace), which is henceforth referred to as ``OSS''.

\begin{table}[!ht]
\centering
\caption{Baseline results}
\label{table:results_baseline}
\resizebox{\columnwidth}{!}{
\begin{tabular}{lrrrrr}
\toprule
  \multirow{2}{*}{\textbf{Recognition system}} & \multirow{2}{*}{\textbf{Workload}} & \multicolumn{3}{c}{\textbf{Open-set}} & \textbf{Closed-set} \\ \cmidrule(r){3-5} \cmidrule(l){6-6} 
  &  &      \textbf{EER} &    $\mathbf{FNIR_{0.1\pmb{\%}}}$ &       \textbf{d'} & \textbf{RR-1} \\
\midrule
    COTS & \multirow{4}{*}{100\%} &   0.011\% &   0.021\% &  8.074 &  100.000\% \\
 ArcFace & &   0.151\% &   0.174\%  &  5.275 &   99.956\%\\
 VGGFace2 & &   5.226\% &  23.066\%  &  2.816 &   96.287\%\\
 FaceNet & &  13.752\% &  46.790\%  &  2.193 &   88.035\% \\
\bottomrule
\end{tabular}
}
\end{table}

\subsection{Selecting Morphing Tools}
\label{subsec:results_selectingmorphing}
A prerequisite for the proposed method to work is that \textit{all} subjects contributing to a morphed image can be successfully matched against it. In other words, the amount of biometric information contained in the morph should be approximately equal for each contributing subject. To check if a morphing tool fulfils this prerequisite, following experiment is conducted: a morph is created from two parent images (references) from two randomly selected subjects. Then, other images (probes) from those two subjects are compared to the created morph. The process is repeated for all data subjects and two score distributions are plotted for the morphing tool described in subsection \ref{subsec:experimentalsetup_morphingtools} and face recognition systems chosen in subsection \ref{subsec:results_selectingrecognition}.

\begin{figure}[!ht]
\centering
\subfloat[OSS face recognition]{
\setlength{\tabcolsep}{0pt}
\begin{tabular}{cc}
\includegraphics[width=0.485\columnwidth]{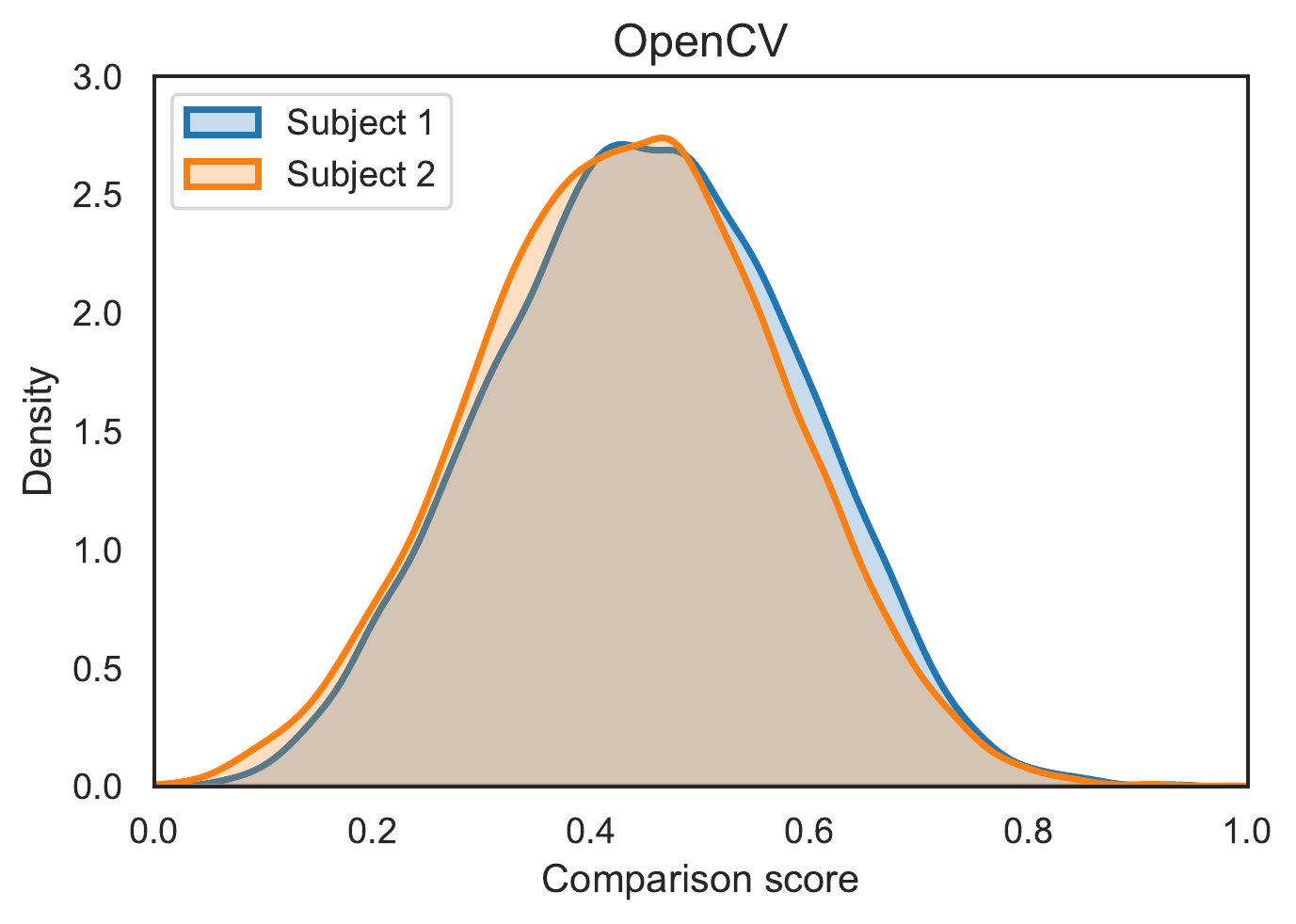} &
\includegraphics[width=0.485\columnwidth]{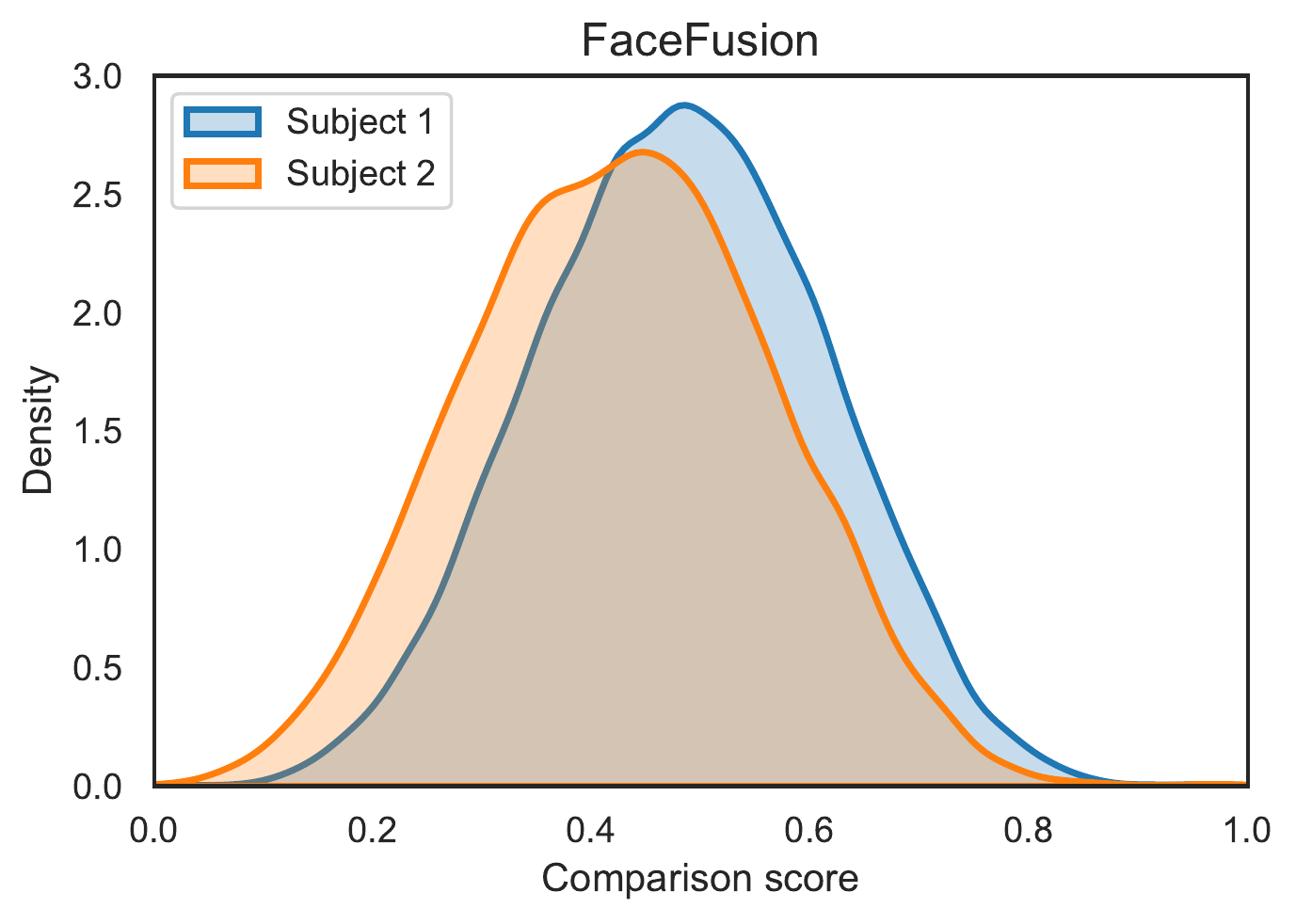} \\
\includegraphics[width=0.485\columnwidth]{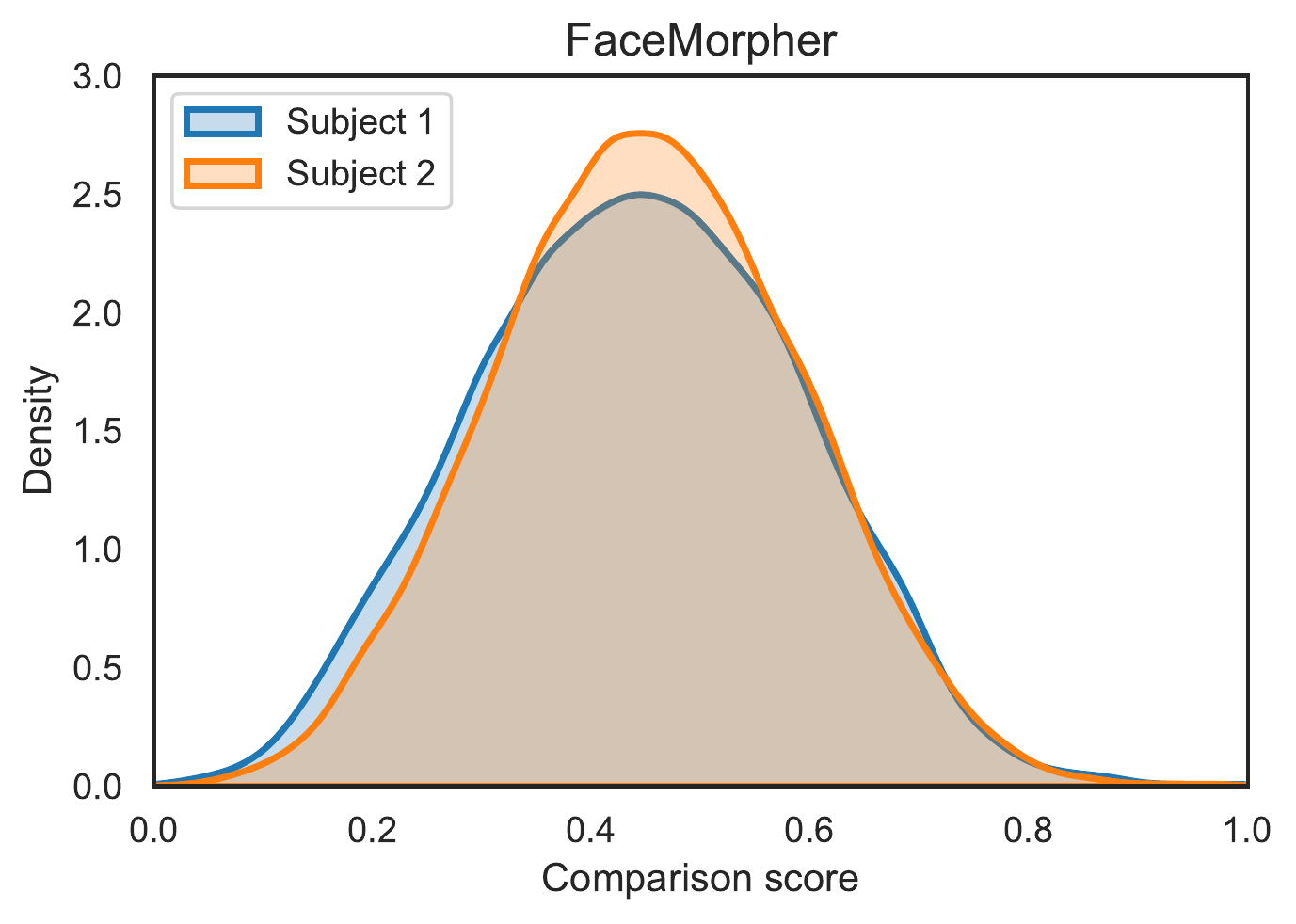} &
\includegraphics[width=0.485\columnwidth]{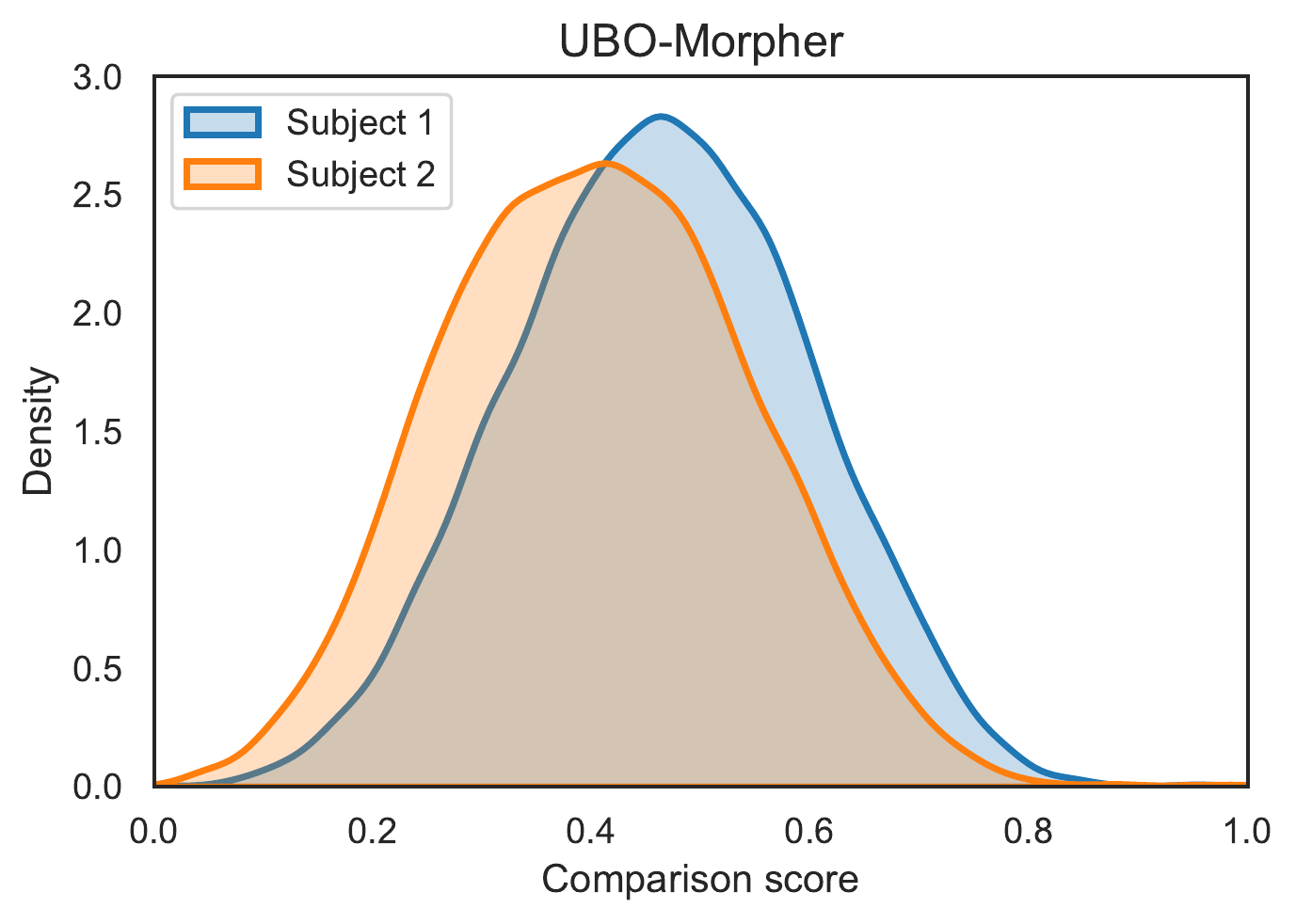}
\end{tabular}
} \\
\subfloat[COTS face recognition]{
\setlength{\tabcolsep}{0pt}
\begin{tabular}{cc}
\includegraphics[width=0.485\columnwidth]{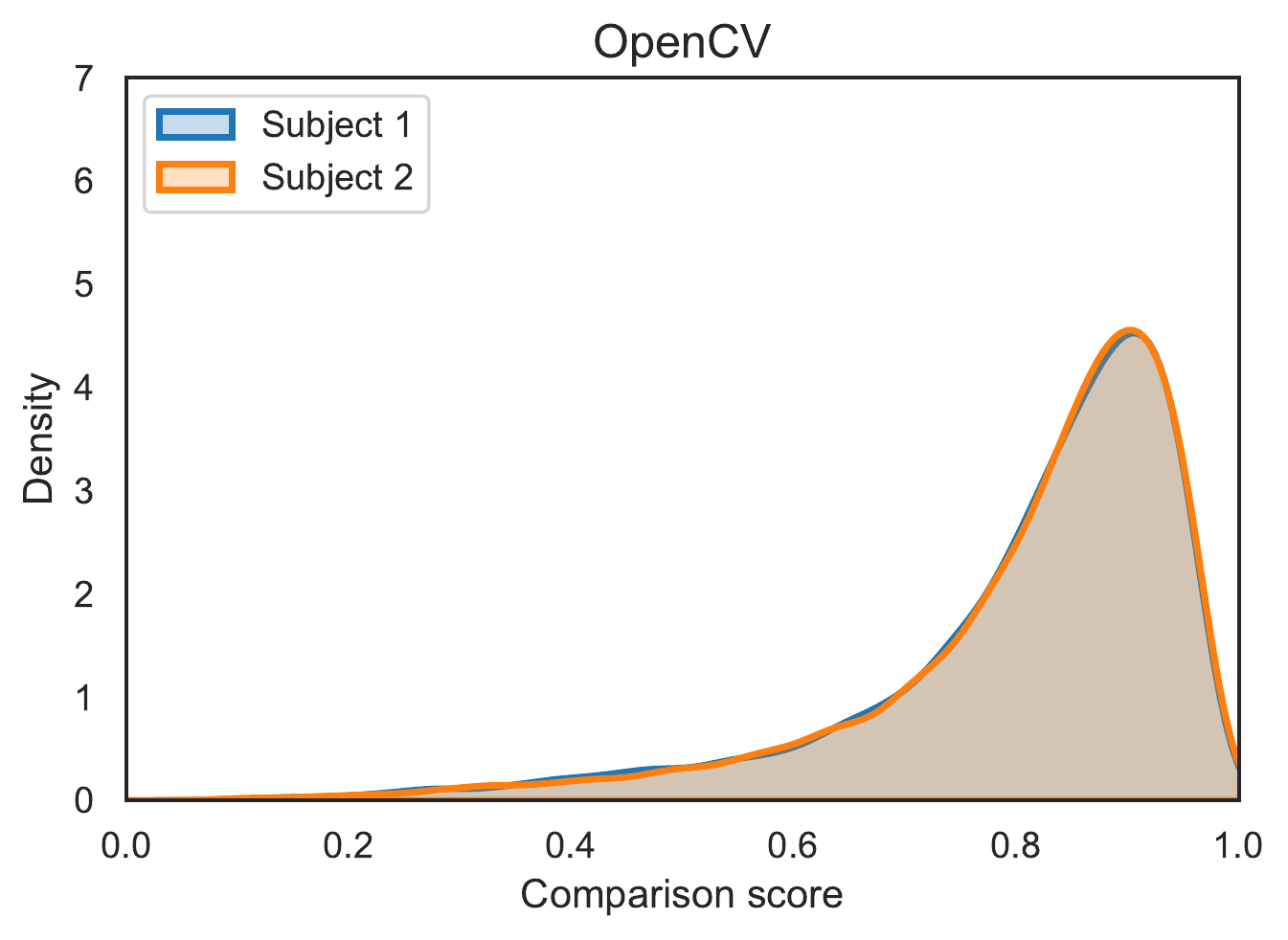} &
\includegraphics[width=0.485\columnwidth]{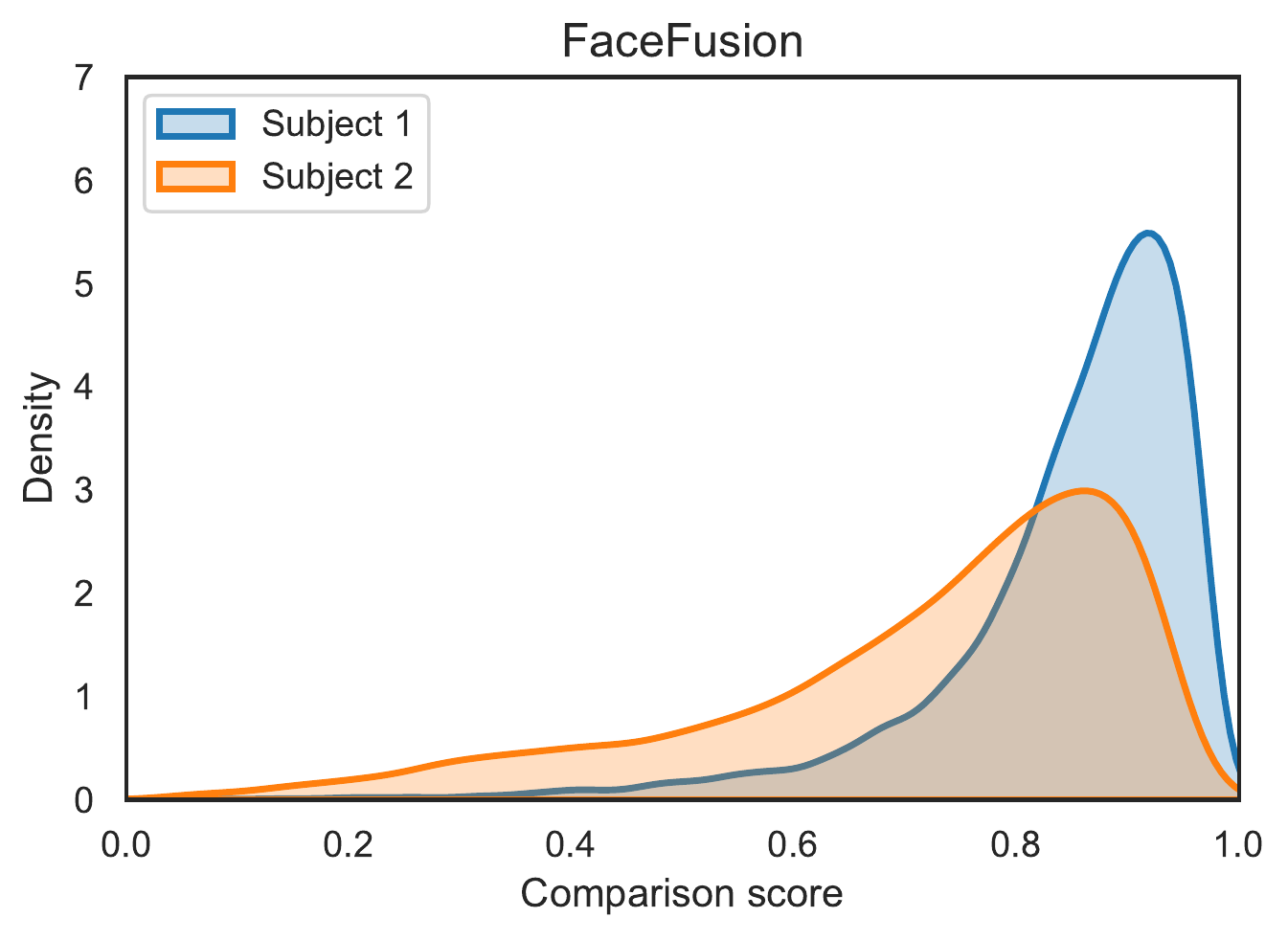} \\
\includegraphics[width=0.485\columnwidth]{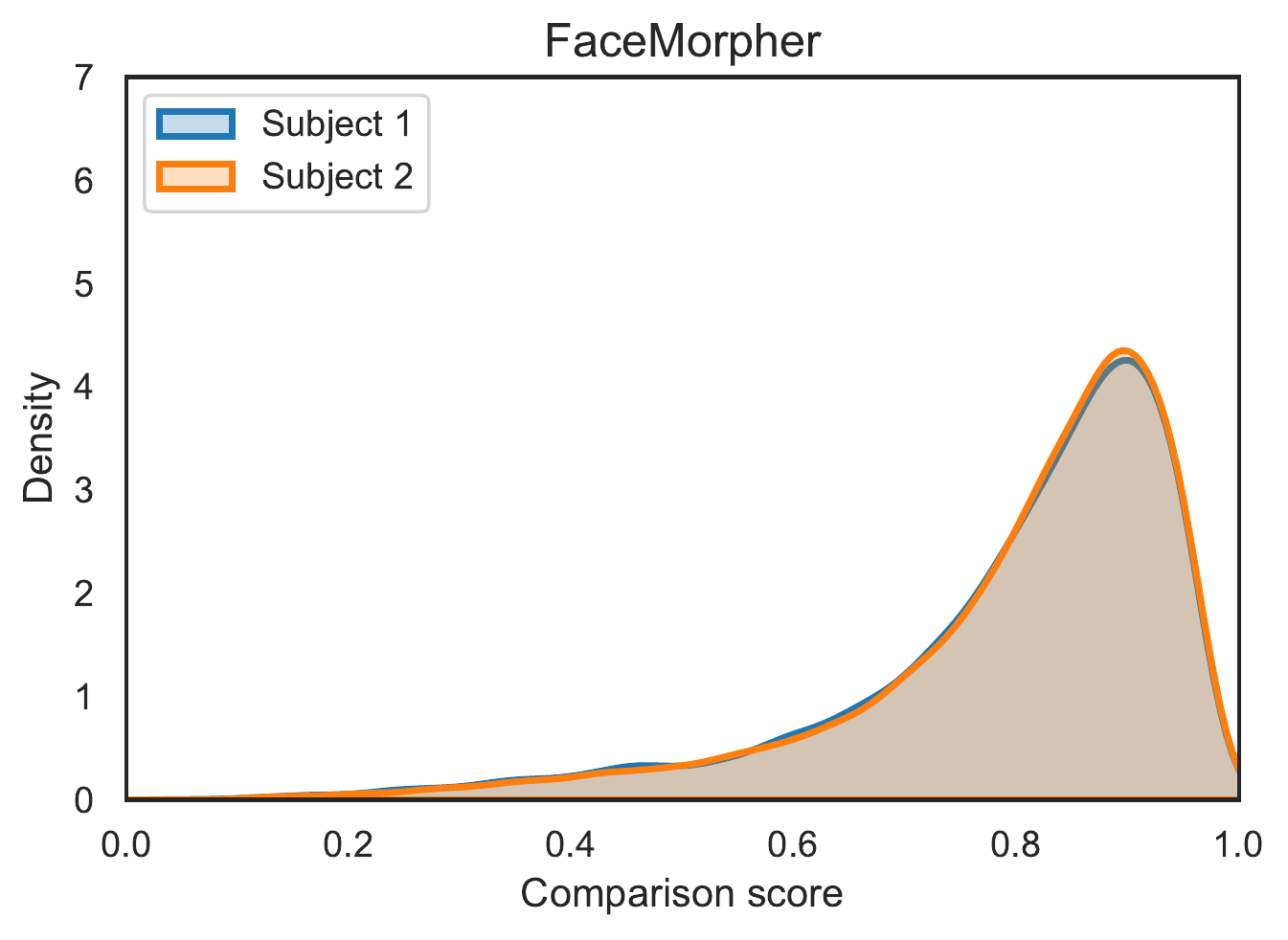} &
\includegraphics[width=0.485\columnwidth]{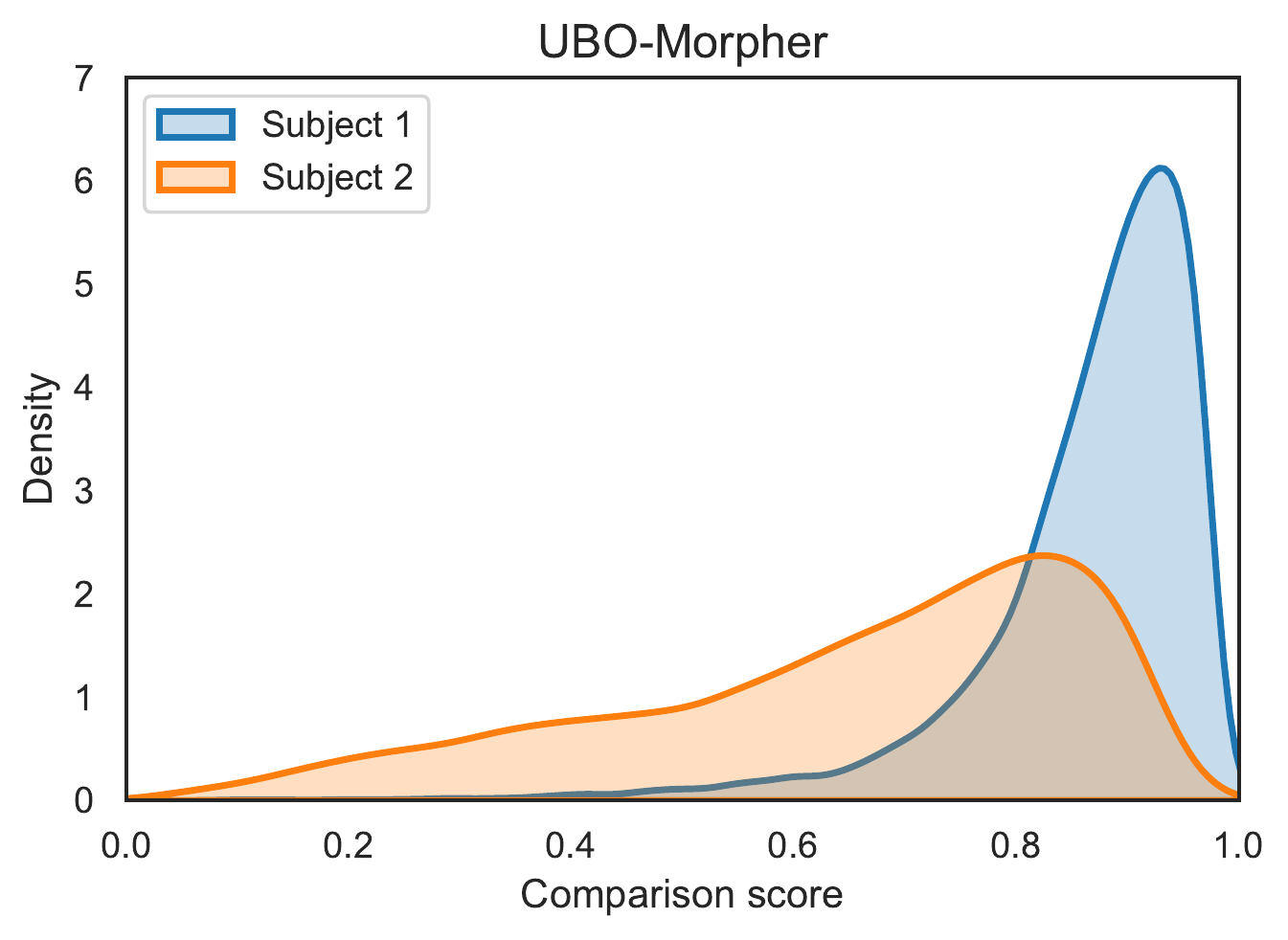}
\end{tabular}
}
\caption{KDEs of comparison score distributions between a morph and subjects contributing to it. Note the divergences between the distributions for some of the morphing tools.}
\label{fig:results_morph_a_b}
\end{figure}

Figure \ref{fig:results_morph_a_b} shows the aforementioned score distributions. It can be seen that for the two tools depicted on the right side of the figure (FaceFusion and UBO-Morpher), the distributions for the two contributing subjects strongly diverge, especially for the COTS system. That is, the second subject tends to receive much worse comparison scores against the morphed image they contributed to; consequently, false negative errors would be more likely to occur for this subject in the context of the proposed multi-step system. The reason for this score divergence presumably lies in the specifics of the algorithm with which the images were morphed. More specifically, in order to avoid artefacts around the head and in the hair region, some tools only conduct morphing within the face region, whereas the surrounding area is cut in from only one (first) of the subjects (\cf figure \ref{fig:example_morphs}). Consequently, such morphs actually contain more biometric information from one of the subjects, which is then reflected in the differences in the comparison scores. 

The distributions depicted on the left side of the figure (FaceMorpher and OpenCV) overlap perfectly for the COTS system; for the OSS system, the overlap is very tight as well, albeit slightly favouring the OpenCV method. To quantitatively confirm those observations, comparisons using the Wasserstein distance are performed. This distance metric is obtained by computing $\int_{-\infty}^{\infty}\left| S_{1} - S_{2} \right|$, where $S_{1}$ and $S_{2}$ denote the cumulative distribution functions for the data displayed in each of the plots in figure \ref{fig:results_morph_a_b}. The results of this evaluation are shown in table \ref{table:results_ws}. The lowest distances between the distributions are observed for the OpenCV morphing tool for both OSS and COTS recognition systems. Based on those quantitative and qualitative results, the OpenCV tool was used for the experiments in the subsequent subsections.

\begin{table}[!ht]
\centering
\caption{Wasserstein distances between the distributions depicted in figure \protect\ref{fig:results_morph_a_b}}
\label{table:results_ws}
\resizebox{0.75\columnwidth}{!}{
\begin{tabular}{lll}
\toprule
\textbf{Recognition system} & \textbf{Morphing tool} & \textbf{Distance} \\
\midrule
COTS & FaceFusion & 0.13220 \\
& FaceMorpher & 0.00423 \\
& OpenCV & 0.00272 \\
& UBO-Morpher & 0.21741 \\
OSS & FaceFusion & 0.05458 \\
& FaceMorpher & 0.01596 \\
& OpenCV & 0.01016 \\
& UBO-Morpher & 0.05785 \\
\bottomrule
\end{tabular}
}
\end{table}

\subsection{Impact of Pairing Algorithm and Morph Capacity}
\label{subsec:results_impactpairing}
\begin{figure}[!ht]
\centering
\subfloat[OSS, random pairing, different morph capacities]{\includegraphics[width=0.485\columnwidth]{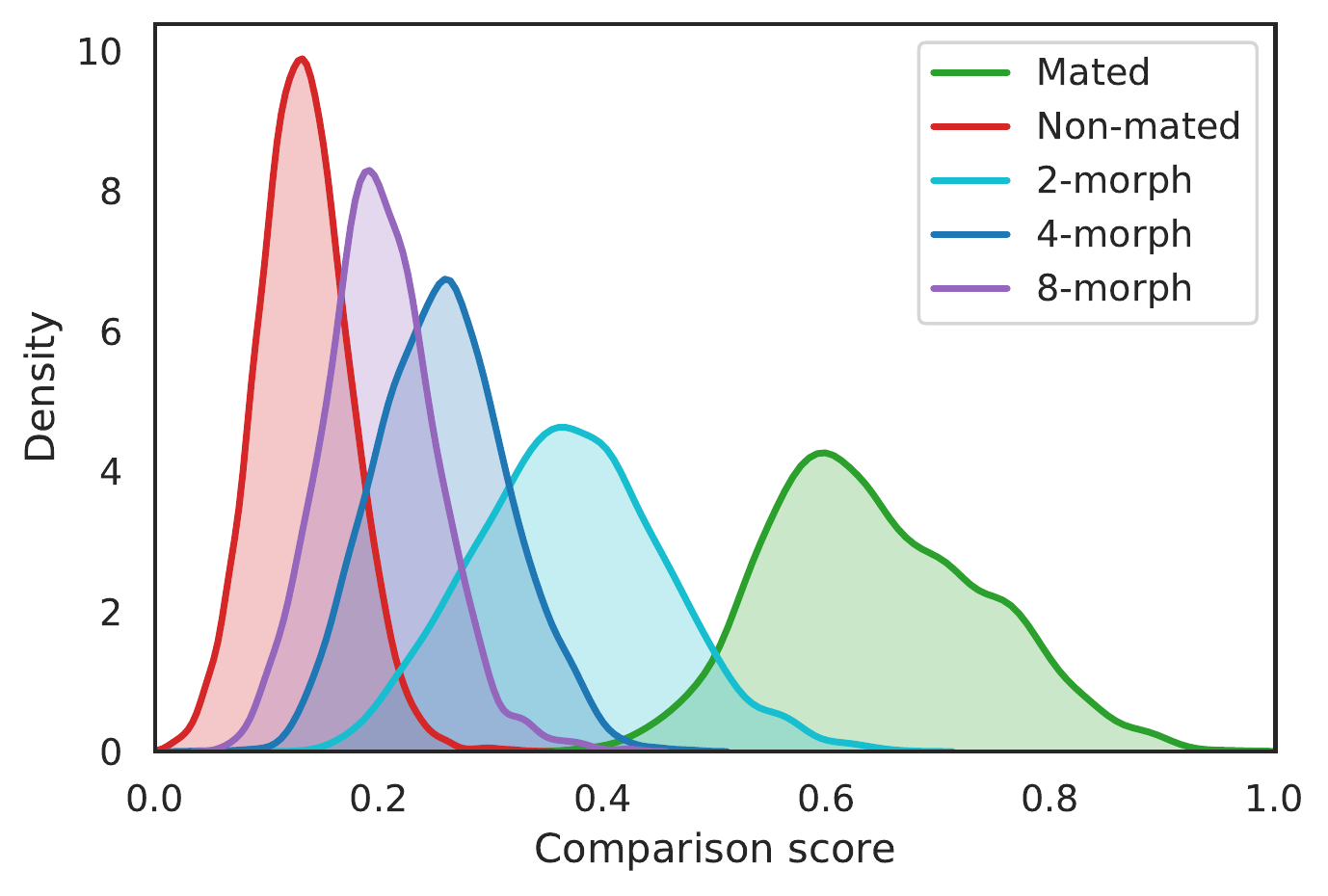}} \hfill
\subfloat[COTS, random pairing, different morph capacities]{\includegraphics[width=0.485\columnwidth]{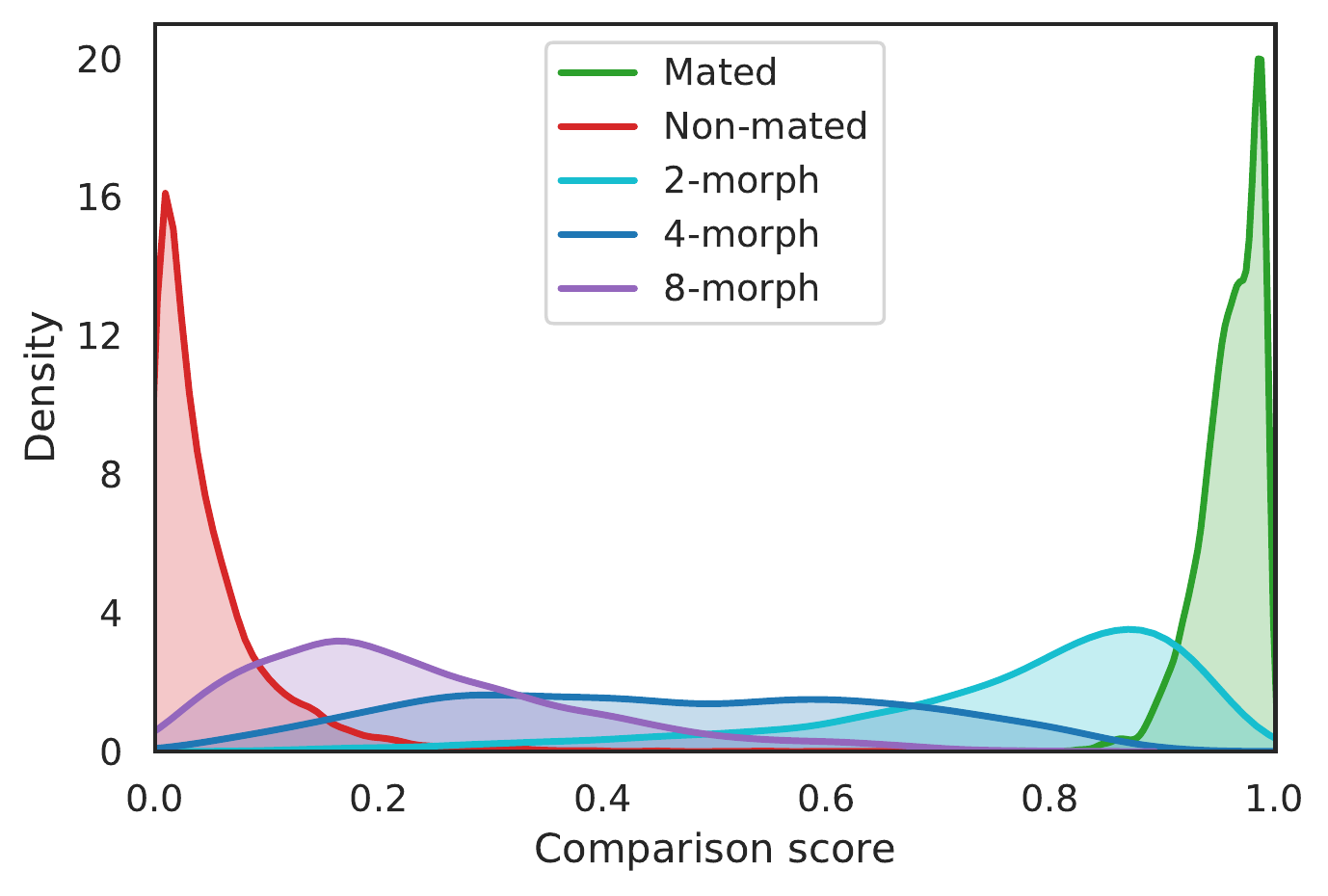}} \\
\subfloat[OSS, intelligent pairing, only 2-morphs]{\includegraphics[width=0.485\columnwidth]{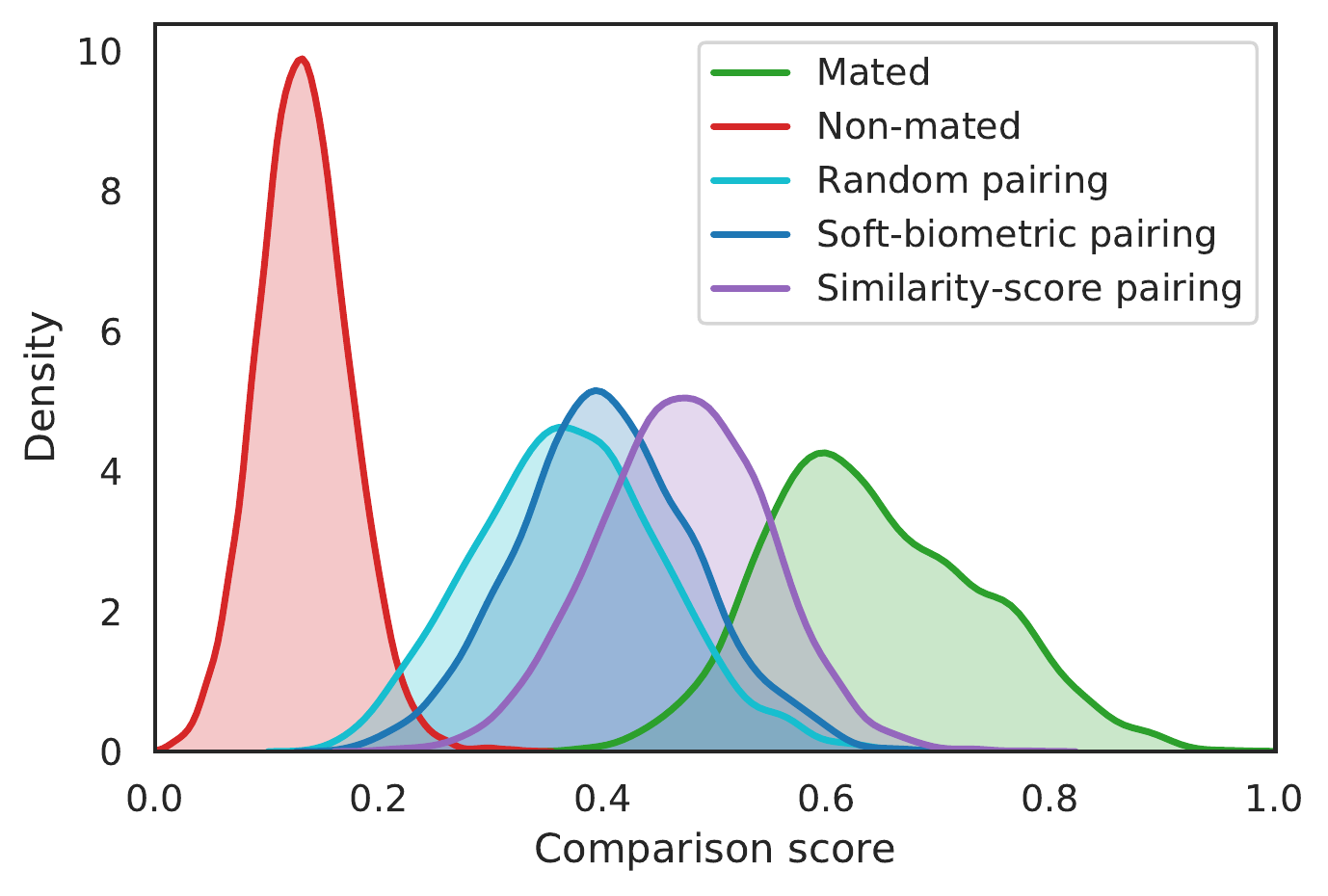}} \hfill
\subfloat[COTS, intelligent pairing, only 2-morphs]{\includegraphics[width=0.485\columnwidth]{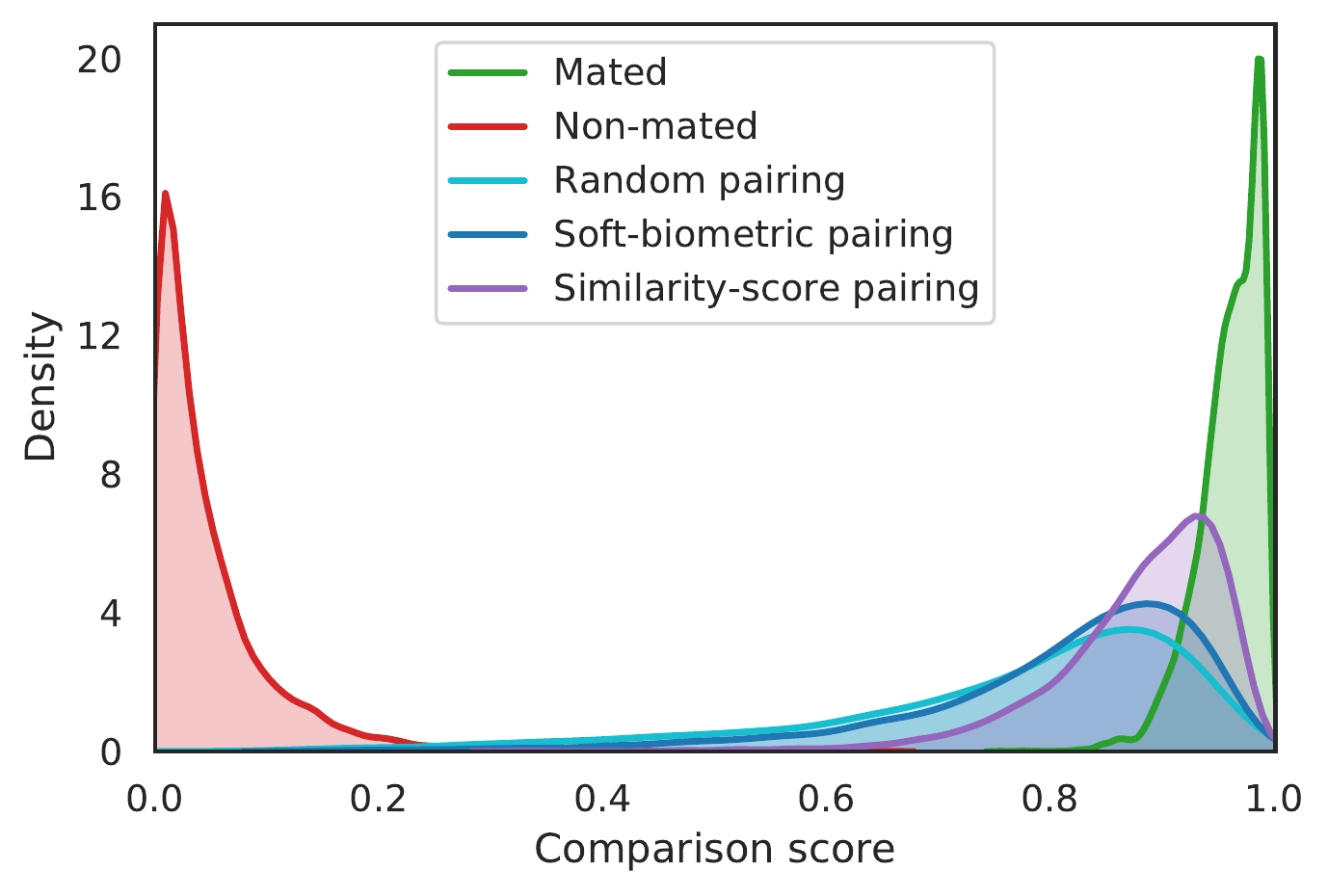}}
\caption{KDEs of comparison score distributions}
\label{fig:results_kde}
\end{figure}

The choice of the image pairing method and the number of subjects contributing to a morph has a non-trivial impact on the system. As demonstrated in figure \ref{fig:results_kde}, by intelligently selecting the pairs, the score distribution for mated-morph comparisons moves towards the distribution of mated scores. The intuitive soft-biometric method performs slightly better than random selection of pairs, whereas significant improvements are observed for the method based on similarity-scores. The aforementioned figure also shows the obvious effects of increasing the capacity of the morphs -- as more subjects are morphed into one image, it tends towards an average face, thereby decreasing the discriminative power. For the tested face recognition systems, it appears that morph capacities beyond $8$ would unlikely be feasible for use in the proposed multi-step retrieval.

\begin{figure*}[!ht]
\centering
\subfloat[Random pairing (with 95\% CI)]{\includegraphics[width=0.325\textwidth]{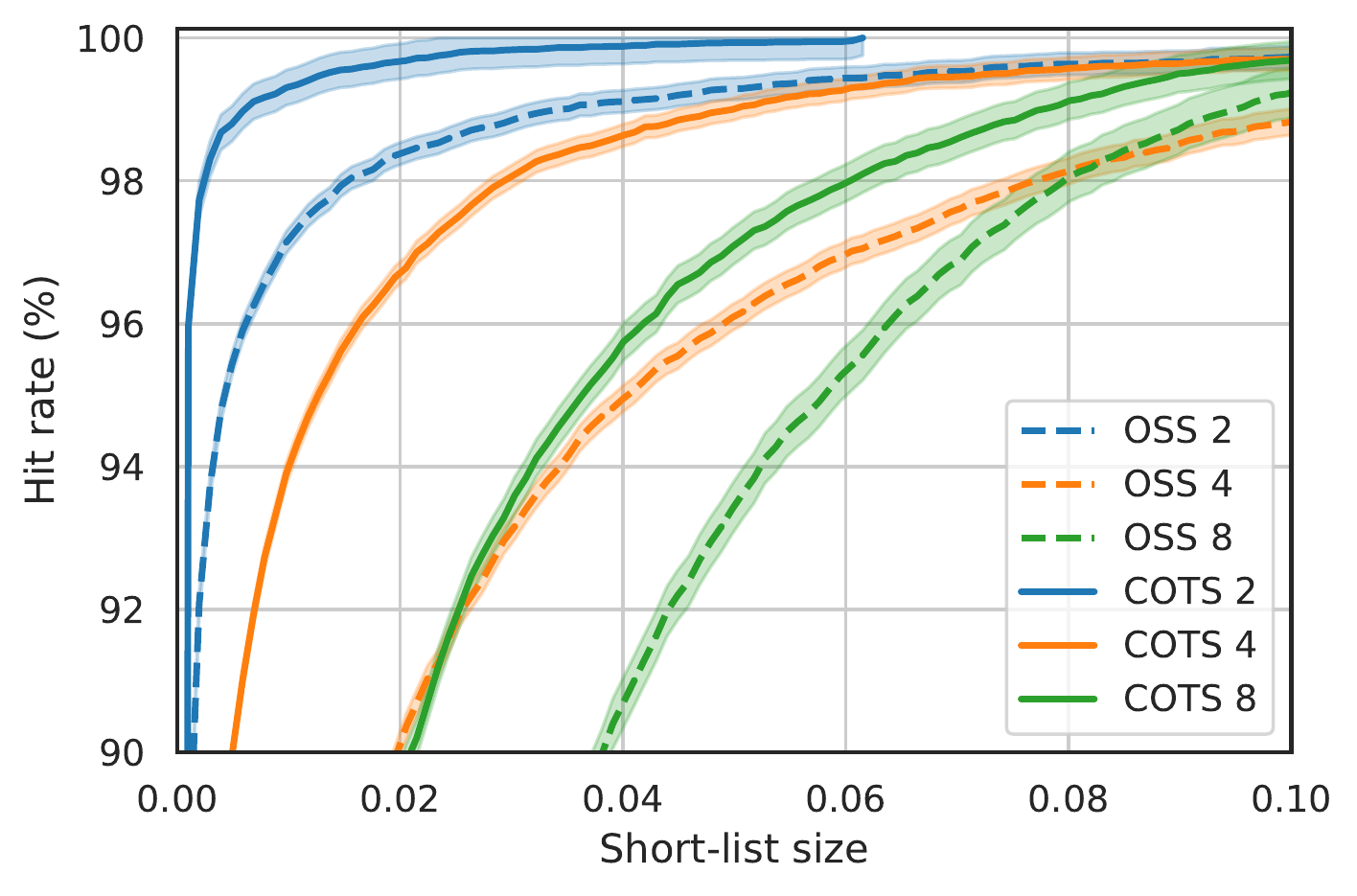}} \hfill
\subfloat[Soft-biometric pairing]{\includegraphics[width=0.325\textwidth]{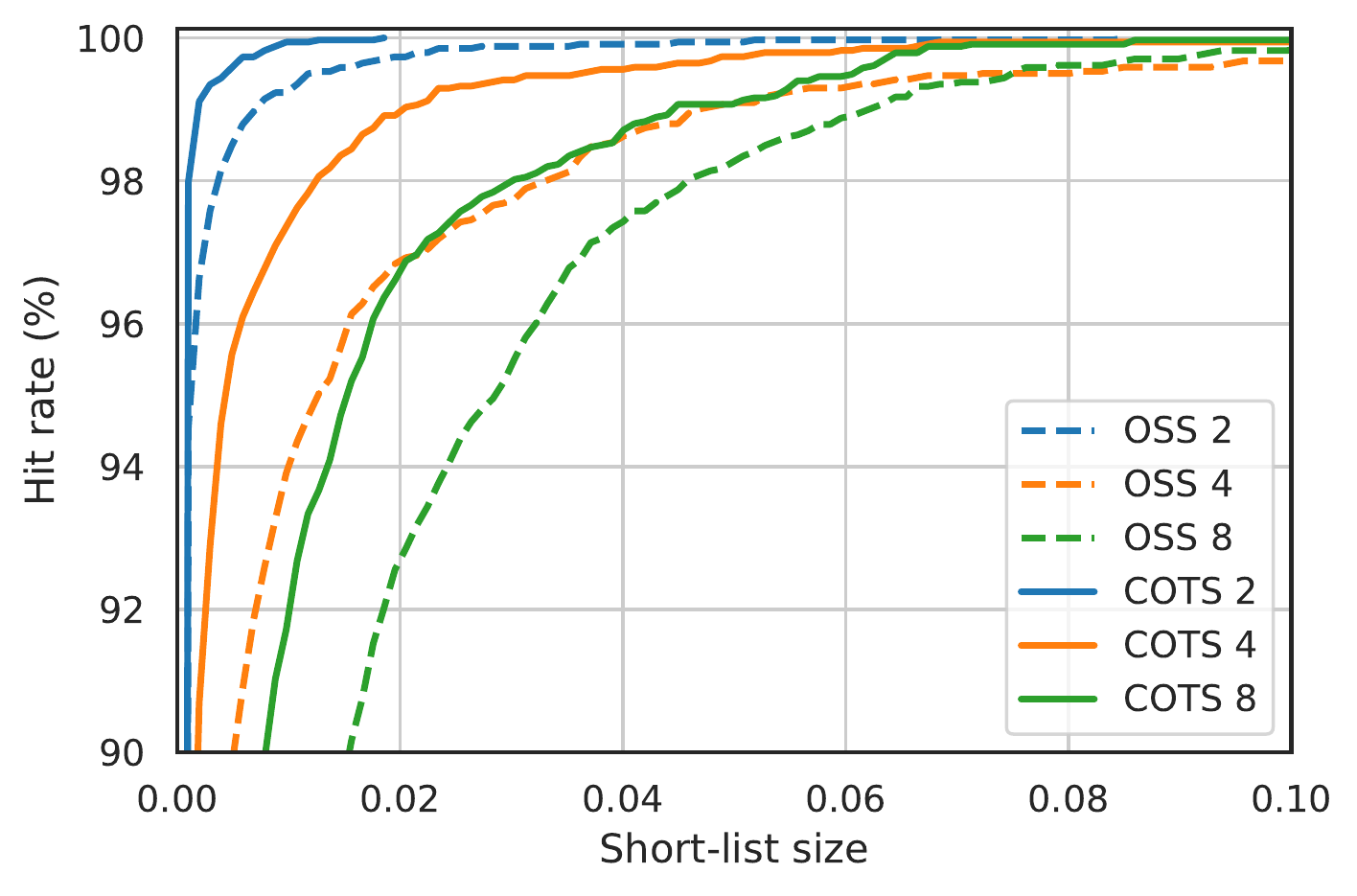}} \hfill
\subfloat[Similarity-score pairing]{\includegraphics[width=0.325\textwidth]{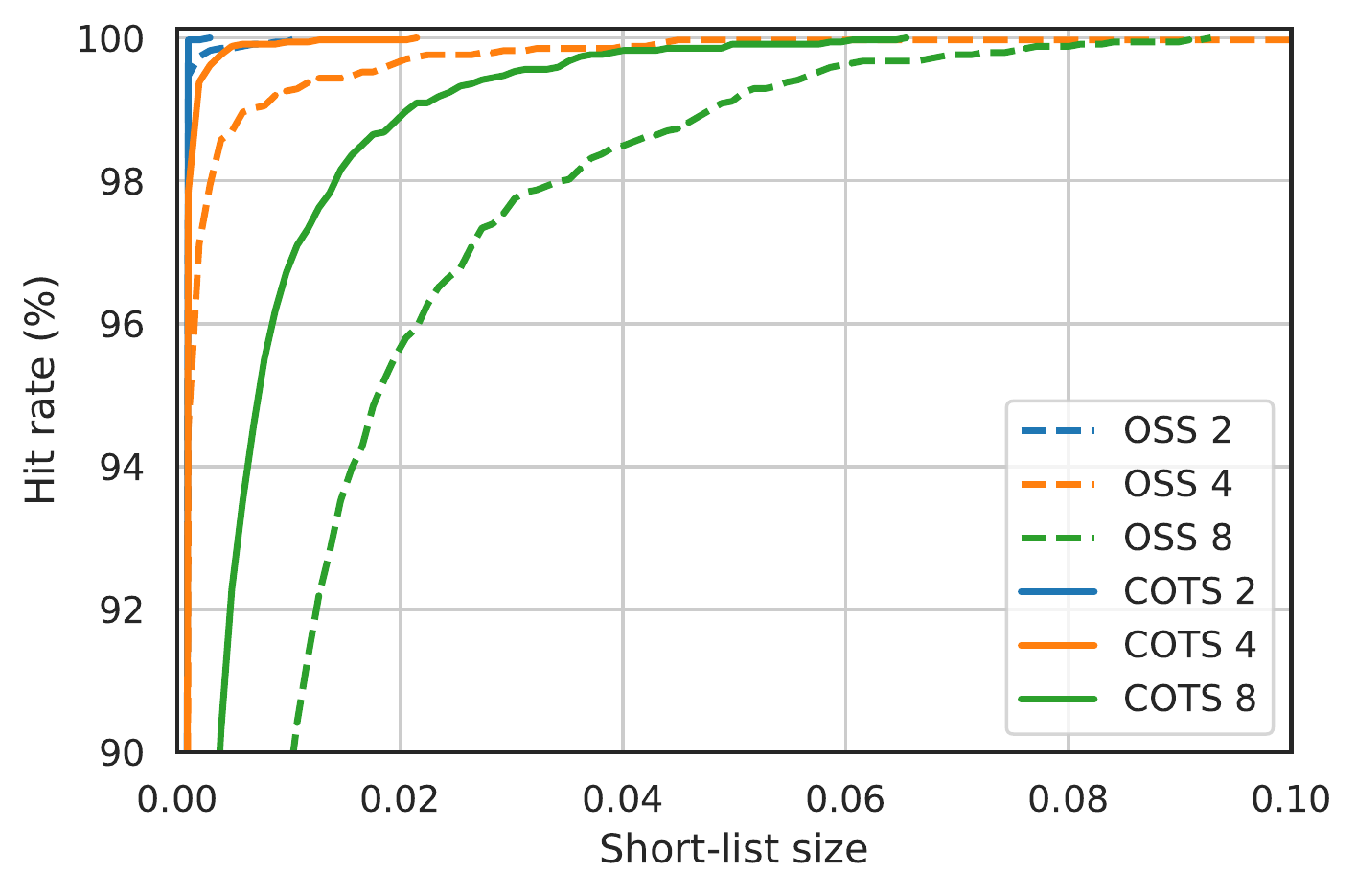}}
\caption{CMC curves}
\label{fig:results_cmc}
\end{figure*}

\begin{table*}[!ht]
\centering
\caption{Summary of the pre-selection results (best result for each HR level for COTS and OSS face recognition marked in \textbf{bold} typeface)}
\label{table:results_hitrate}
\resizebox{\textwidth}{!}{
\scriptsize
\begin{tabular}{lllllll}
	\toprule
	\multirow{2}{*}{\textbf{Morph size}} &
	\multirow{2}{*}{\textbf{Recognition system}} &
	\multirow{2}{*}{\textbf{Pairing algorithm}} &
	\multicolumn{4}{c}{\textbf{Workload at}} \\ \cmidrule{4-7}
	& & & 95\% HR & 99\% HR & 99.5\% HR & 100\% HR\\
	\midrule
	2 & COTS & Random & 50.20\% $\pm$ 0.00\%  & 51.29\% $\pm$ 0.22\%  & 52.85\% $\pm$ 0.37\%  & 72.73\% $\pm$ 14.62\% \\
	 &  & Soft-biometric & 50.20\%  & 50.39\%  & 50.98\%  & 53.71\% \\
	 & & Similarity-score & 50.20\%  & 50.20\%  & 50.20\%  & 50.59\% \\
	 & OSS & Random & 50.90\% $\pm$ 0.13\%  & 57.11\% $\pm$ 2.31\%  & 63.79\% $\pm$ 4.63\%  & 98.32\% $\pm$ 19.68\% \\
	 &  & Soft-biometric & 50.39\%  & 51.56\%  & 52.54\%  & 66.99\% \\
	 &  & Similarity-score & 50.20\%  & 50.20\%  & 50.39\%  & \textbf{52.15\%} \\
	4 & COTS & Random & 30.31\% $\pm$ 0.74\%  & 45.16\% $\pm$ 3.93\%  & 55.08\% $\pm$ 6.07\%  & 108.36\% $\pm$ 11.84\% \\
	 &  & Soft-biometric & 26.95\%  & 33.20\%  & 39.45\%  & 80.08\% \\
	 &  & Similarity-score & 25.39\%  & \textbf{25.78\%}  & \textbf{26.17\%}  & \textbf{33.59\%} \\
	 & OSS & Random & 41.48\% $\pm$ 1.89\%  & 67.42\% $\pm$ 3.35\%  & 78.98\% $\pm$ 6.29\%  & 117.73\% $\pm$ 10.37\% \\
	 &  & Soft-biometric & 30.08\%  & 43.75\%  & 53.91\%  & 121.09\% \\
	 &  & Similarity-score & \textbf{25.78\%}  & \textbf{27.73\%}  & \textbf{31.64\%}  & 77.34\% \\
	8 & COTS & Random & 42.03\% $\pm$ 1.26\%  & 75.47\% $\pm$ 2.79\%  & 85.47\% $\pm$ 3.86\%  & 109.69\% $\pm$ 2.88\% \\
	 &  & Soft-biometric & 25.00\%  & 48.44\%  & 61.72\%  & 95.31\% \\
	 &  & Similarity-score & \textbf{18.75\%}  & 29.69\%  & 36.72\%  & 64.84\% \\
	 & OSS & Random & 58.91\% $\pm$ 2.43\%  & 88.75\% $\pm$ 2.62\%  & 97.66\% $\pm$ 3.07\%  & 111.09\% $\pm$ 3.39\% \\
	 &  & Soft-biometric & 35.94\%  & 62.50\%  & 72.66\%  & 112.5\% \\
	 & & Similarity-score & 27.34\%  & 51.56\%  & 58.59\%  & 86.72\% \\
	\bottomrule
\end{tabular}

}
\end{table*}

To further evaluate the impact of the two aforementioned factors, closed-set identification experiments were carried out. The resulting CMC curves are shown in figure \ref{fig:results_cmc}. It can be observed that the curves quickly converge onto 100\% HR for morph capacities of $2$ and $4$ subjects; in other words, this means that only very small short-list size would be required to avoid pre-selection errors. For the morph capacity of $8$ subjects, the curves reach the 100\% point much later, \ie larger short-lists would be required. This does not necessarily disqualify configurations with such a morph capacity, as the overall computational workload depends not only on the short-list size of the pre-selected short-list, as described in subsection \ref{subsec:proposedsystem_retrieval}. Furthermore, such morphs could potentially be used in the multi-step system or for applications where some pre-selection errors might be tolerable. This aspect of the trade-off between biometric performance and computational workload is addressed in more detail in the next subsection.
From the figure \ref{fig:results_kde}, the impact of the pairing method is very clear -- vast improvements can be observed when moving from the random pairing to pairing based on soft-biometrics and even more so when the pairing is based on similarity-scores. Both COTS and OSS face recognition systems perform well, with the COTS system being the better of the two, which is to be expected intuitively and based on the baseline results from subsection \ref{subsec:results_selectingrecognition}.

\subsection{Overall Results}
\label{subsec:results_overall}
Table \ref{table:results_hitrate} summarises the results of the pre-selection in a two-step retrieval system for several commonly reported levels of hit rate. Since computational workload is reduced relative to the baseline by using the proposed method (\cf table \ref{table:results_baseline}), it follows that $0\% < \text{W}_{\text{Proposed}} < 100\%$. For $\text{HR}>99\%$, lowest computational workload is achieved with pairing based on similarity-scores and a morph capacity of 4 contributing subjects for the COTS face recognition system. For the OSS face recognition system, its lower discriminative power has to be mitigated by selecting a lower morph capacity of 2. At the stringent 99.5\% or even 100\% HR level, both the OSS and COTS systems approach their respective lower limits (\cf figure \ref{fig:comparisons_example_twostage}) of computational workload for the selected morph capacities, \ie approximately $\frac{1}{2}$ and $\frac{1}{4}$ of the baseline workload.

Table \ref{table:results_proposed} summarises the overall results of the best configuration of the proposed system in open-set and closed-set identification scenarios. For both COTS and OSS face recognition, the biometric performance of the baseline method (\cf table \ref{table:results_baseline}) is maintained, while the proposed system allows for the computational workload to be reduced by approximately 70\% and 50\% for COTS and OSS face recognition systems, respectively. Furthermore, for the COTS system, a slight improvement (lower workload) from two-step to multi-step retrieval method can be observed.

\begin{table*}[!ht]
\centering
\caption{Summary of the proposed system results}
\label{table:results_proposed}
\resizebox{0.75\textwidth}{!}{
\begin{tabular}{llrrrrr}
\toprule
  \multirow{2}{*}{\textbf{Recognition system}} & \multirow{2}{*}{\textbf{Configuration}} & \multirow{2}{*}{\textbf{Workload}} & \multicolumn{3}{c}{\textbf{Open-set}} & \textbf{Closed-set} \\ \cmidrule(r){4-6} \cmidrule(l){7-7} 
  & &  &     \textbf{EER} &    $\mathbf{FNIR_{0.1\pmb{\%}}}$ &       \textbf{d'} & \textbf{RR-1} \\
\midrule
    COTS & Two-step & 33.59\% &  0.029\% &   0.029\% & 8.004 &  100.000\%  \\
         & Multi-step & 29.88\% &  0.029\% &   0.029\% & 7.980 & 100.000\%  \\
    OSS & Two-step & 52.15\% & 0.159\% &   0.264\% & 5.136 &  99.956\%  \\
         & Multi-step & 53.32\% & 0.159\% &   0.268\% & 5.155 & 99.956\%  \\
\bottomrule
\end{tabular}
}
\end{table*}

\section{Discussion}
\label{sec:discussion}
In subsection \ref{subsec:comparison_existing}, the proposed system is compared \wrt other existing methods of computational workload reduction. Subsection \ref{subsec:scalability} discusses the scalability of the proposed system.

\subsection{Comparison with Existing Systems}
\label{subsec:comparison_existing}
While methods with an even better computational workload reduction than the proposed method have appeared in the academic literature (see \cite{Drozdowski-WorkloadSurvey-IET-2019}), they require access to more information or are less flexible than the proposed method. More specifically, such approaches often utilise the feature vectors of the underlying facial recognition system or additional (potentially error-prone) classifiers; furthermore, other prerequisites such as extensive training or simple template comparators may limit their flexibility. Due to this reason, a direct benchmark against such methods would be cumbersome due to different aspects of the system being focused on. More concretely, the proposed method focuses on a balance between computational workload reduction and high flexibility irrespective of the used face recognition system, whereas the related works tend to focus on maximising the computational workload reduction by being strongly interwoven with the underlying biometric recognition system. Consequently, the vast majority of the approaches proposed in the literature cannot be combined with COTS recognition systems, due to their operation as black-box systems.

\subsection{Scalability}
\label{subsec:scalability}
The concepts underlying the proposed indexing scheme can be theoretically seamlessly scaled with the growing size of the enrolment database and can be trivially parallelised or distributed. Furthermore, the proposed retrieval algorithm features a flexible design which facilitates dynamic adjustment of decision thresholds and setting pre-selection subset sizes relative to the enrolment database size. One important consideration \wrt the scalability would be the increased computational costs of the pairing algorithm (subsection \ref{subsec:proposedsystem_pairselection}); those could, however, be mitigated by computation distribution or additional bucketing of the enrolment database. On the other hand, a larger size of the enrolment database would likely increase the probability of finding suitable pairings for all the subjects (especially for the outliers). This in turn would be expected to improve the quality of the morphing process and the retrieval algorithm. Thus, the proposed methods can be expected to scale both in terms of biometric recognition performance and computational efficiency.

As mentioned in subsection \ref{subsec:experimentalsetup_datasets}, the proposed method requires somewhat constrained data of certain quality. While directly testing the scalability of the proposed method with a larger dataset would certainly be of interest, currently no large-scale, publicly available datasets of facial images with sufficient image quality exist. Images in well-known large-scale datasets, \eg LFW \cite{Huang-LFW-2007} or MegaFace \cite{Kemelmacher-MegaFace-2016}, do not generally possess a sufficient image quality for the proposed method. While this can certainly be considered a limitation of the proposed method, it is worth noting that many existing biometric systems, \eg border control, operate with images of high-quality, thereby making the application of the proposed method theoretically feasible (see section \ref{sec:conclusion} for more details).

\section{Conclusion}
\label{sec:conclusion}
In this article, facial image morphing, which constitutes a serious attack vector against biometric systems has been reconceptualised for use in a beneficial manner. Specifically, a method of indexing biometric data with signal-level fusion has been presented. The proposed method relies on intelligent pairing and morphing of facial parent images to facilitate a multi-step retrieval for biometric identification transactions. In a comprehensive experimental evaluation with open-source and commercial systems, the proposed method has been shown to achieve a biometric performance nearly identical to that of an exhaustive search-based baseline, while simultaneously substantially reducing the computational workload of biometric identification transactions (down to ${\sim}30\%$ at 100\% HR). 

In contrast to related works, the proposed method could be effortlessly integrated even into black-box biometric recognition systems, as it merely requires access to the raw biometric samples (facial images) and the comparison scores. One limitation of the proposed method is the requirement of good-quality reference images. This notwithstanding, numerous operational systems could benefit of the proposed method, as they store or process images compliant with very strict quality standards \cite{ICAO-9303-p9-2015}, \eg biometric samples used in passports or visa applications. Furthermore, the enrolment of new subjects into an existing system would require a periodical re-computation of the index.

Although out of scope for this article, the developed concepts may also be applied and benchmarked in conjunction with a feature-level fusion in the context of white-box systems, which would constitute an interesting item of future research.

\section*{Acknowledgements}
\label{sec:acknowledgements}
This research work has been funded by the German Federal Ministry of Education and Research and the Hessian Ministry of Higher Education, Research, Science and the Arts within their joint support of the National Research Center for Applied Cybersecurity ATHENE.
\balance
\bibliographystyle{IEEEtran}
\bibliography{references}

\end{document}